\documentclass[letterpaper, 10 pt, conference]{ieeeconf}  % Comment this line out if you need a4paper

\IEEEoverridecommandlockouts                              % This command is only needed if
\usepackage{amsmath}
\usepackage{amssymb}
\usepackage{graphicx}
\usepackage{array}
\usepackage{booktabs}
\usepackage{multirow}
\usepackage[table,xcdraw]{xcolor}
\usepackage{cite}
\usepackage{makecell}
\usepackage{threeparttable}
\usepackage{subfig}
\usepackage{xcolor}
\usepackage{url}
\usepackage[hidelinks]{hyperref}

\title{\LARGE \bf
If It Moves, Radar Knows: A Physics-Aware Radar Transformer for
Class-Agnostic Moving-Object Detection
}

\author{Yinghao Sun, Shuguang Li$^*$, Jinliang Shao and Tieshan Li% <-this % stops a space
\thanks{The authors are all with School of Automation Engineering of the University of Electronic Science and Technology of China (UESTC), Chengdu, Sichuan, P. R. China {\tt\small sunyinghao@uestc.edu.cn, lisg@uestc.edu.cn, jinliangshao@uestc.edu.cn, litieshan073@uestc.edu.cn}}% <-this % stops a space
}

\begin{document}

\maketitle
\thispagestyle{empty}
\pagestyle{empty}

%%%%%%%%%%%%%%%%%%%%%%%%%%%%%%%%%%%%%%%%%%%%%%%%%%%%%%%%%%%%%%%%%%%%%%%%%%%%%%%%
\begin{abstract}

Detectors trained on closed-set annotations can miss rare moving objects
outside the training taxonomy. Automotive radar provides
category-independent motion evidence through Doppler measurements and is less
affected by adverse illumination and weather, but its sparse and noisy returns
hinder conventional class-aware 3D box detection. For downstream motion reasoning and collision avoidance, surface
location and velocity remain directly useful even when complete box geometry
is difficult to recover. We present the Physics-Aware Radar Transformer (PART), a fully sparse
radar-only detector that predicts an existence confidence, a representative
surface point, and 2D ground-plane velocity for each moving-object hypothesis.
Doppler-Aware Query Initialization (DAQI) replaces scene-independent learned
queries with input-dependent proposals obtained by clustering radar returns
jointly in position and velocity, easing query--object assignment in sparse
radar scenes. Physics-Guided Cross-Attention (PGCA) incorporates
radial-Doppler consistency and radar cross section (RCS) information into
query--point association. Uncertainty-aware supervision randomly masks
ground-truth objects and assigns soft existence targets to ambiguous
radar-supported queries, reducing reliance on exhaustive annotations. With only 1.1 million parameters, PART achieves a
class-agnostic average precision (CA-AP) of $0.8827$, a mean average surface
translation error (mASTE) of $0.3188$~m, and a mean average velocity error
(mAVE) of $0.8084$~m/s on nuScenes. It further attains $0.9203$ recall on rare and safety-relevant categories
excluded from the standard evaluation and remains effective at night, in
rain, and under severe occlusion. Inspection of apparent false positives
further shows that some PART predictions correspond to moving objects absent
from the nuScenes annotations. Code and pretrained model weights will be publicly available at \url{https://github.com/sunyinghao-uestc/PART}.

\end{abstract}

\begin{figure*}[!t]
\centering
\subfloat[Powered wheelchair user in profile.]{
    \includegraphics[width=0.24\linewidth]{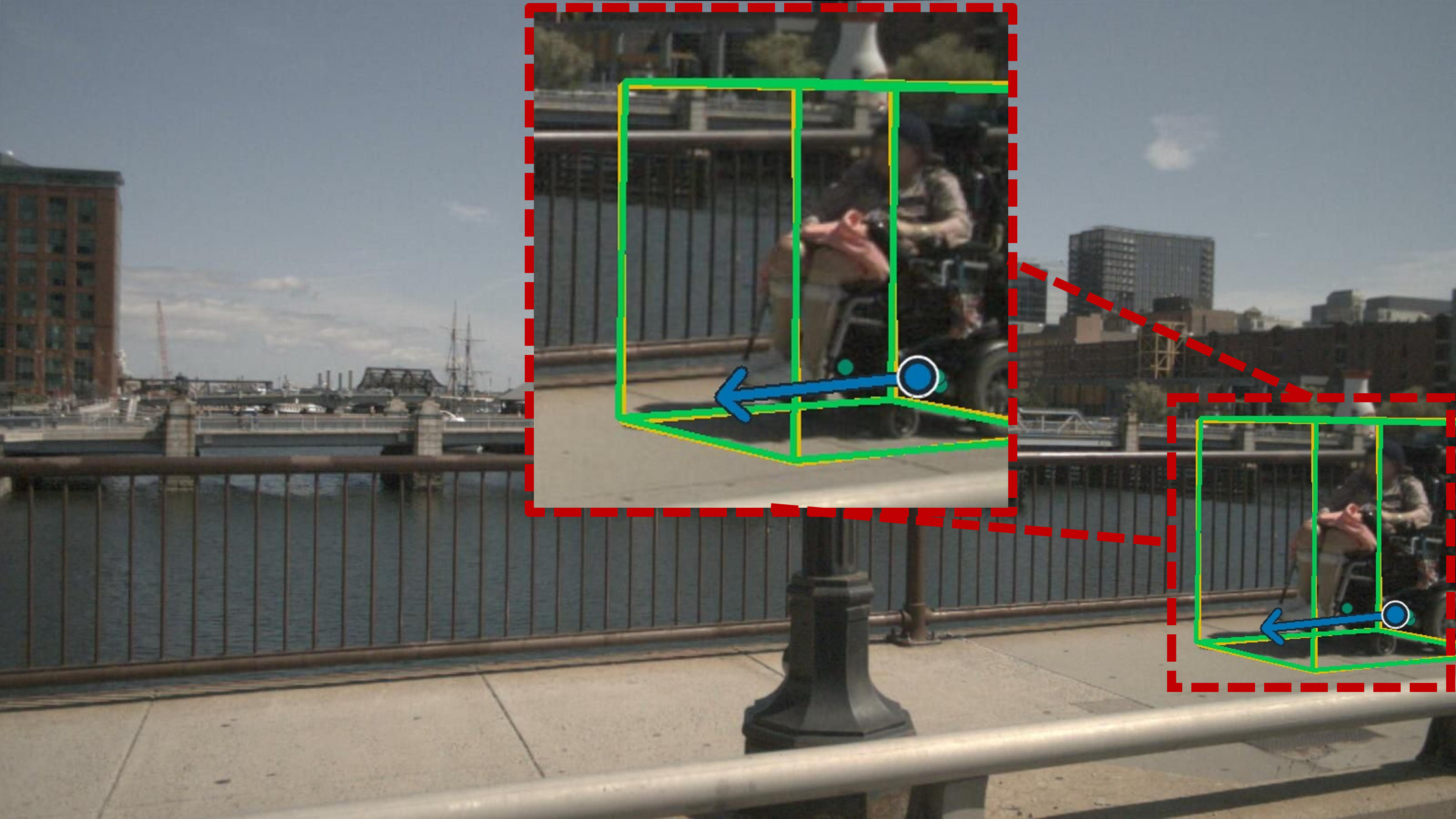}}
\hfill
\subfloat[Powered wheelchair user, frontal view.]{
    \includegraphics[width=0.24\linewidth]{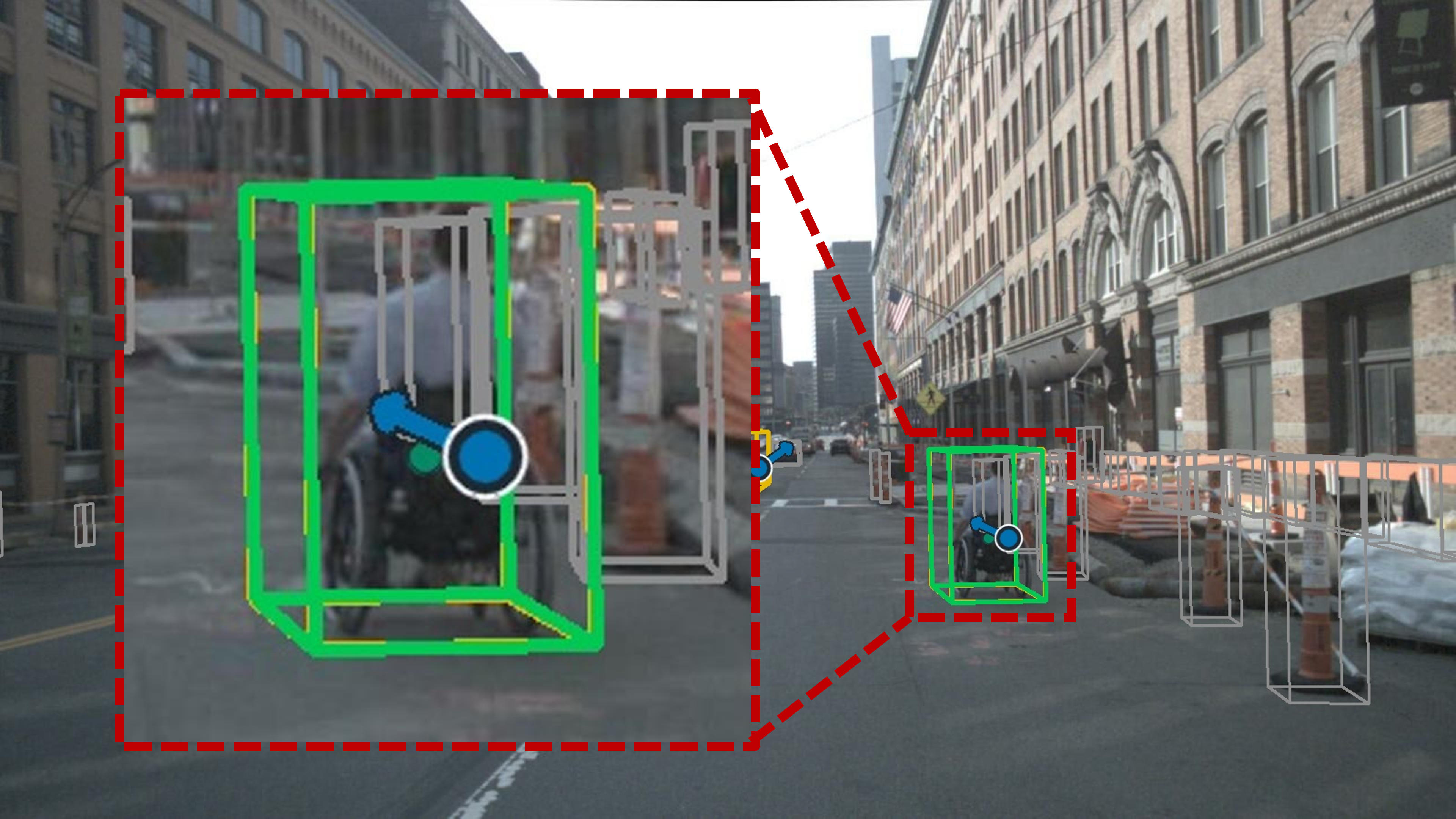}}
\hfill
\subfloat[Personal-mobility rider behind a guardrail.]{
    \includegraphics[width=0.24\linewidth]{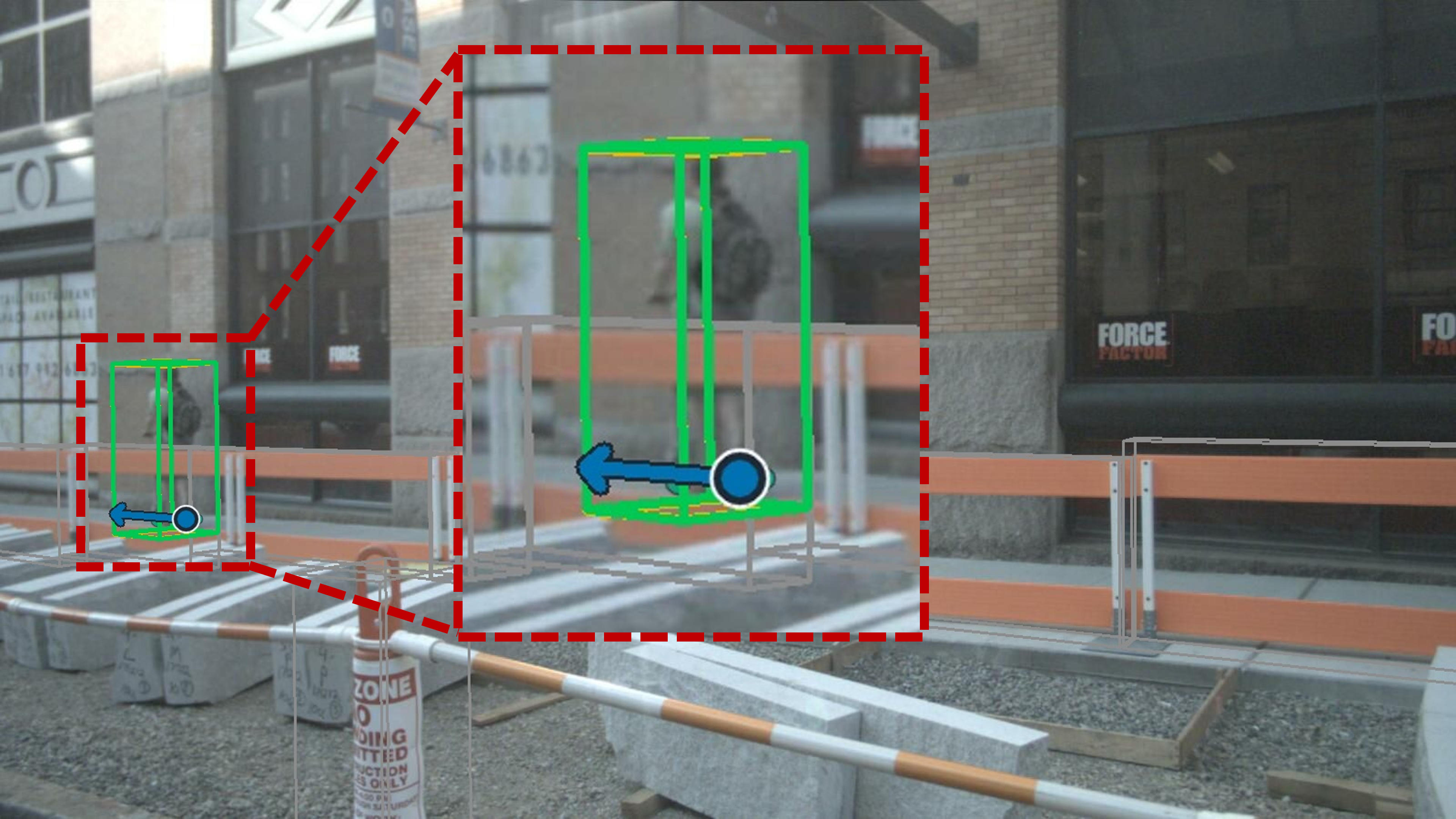}}
\hfill
\subfloat[Dog-walking pedestrian at night.]{
    \includegraphics[width=0.24\linewidth]{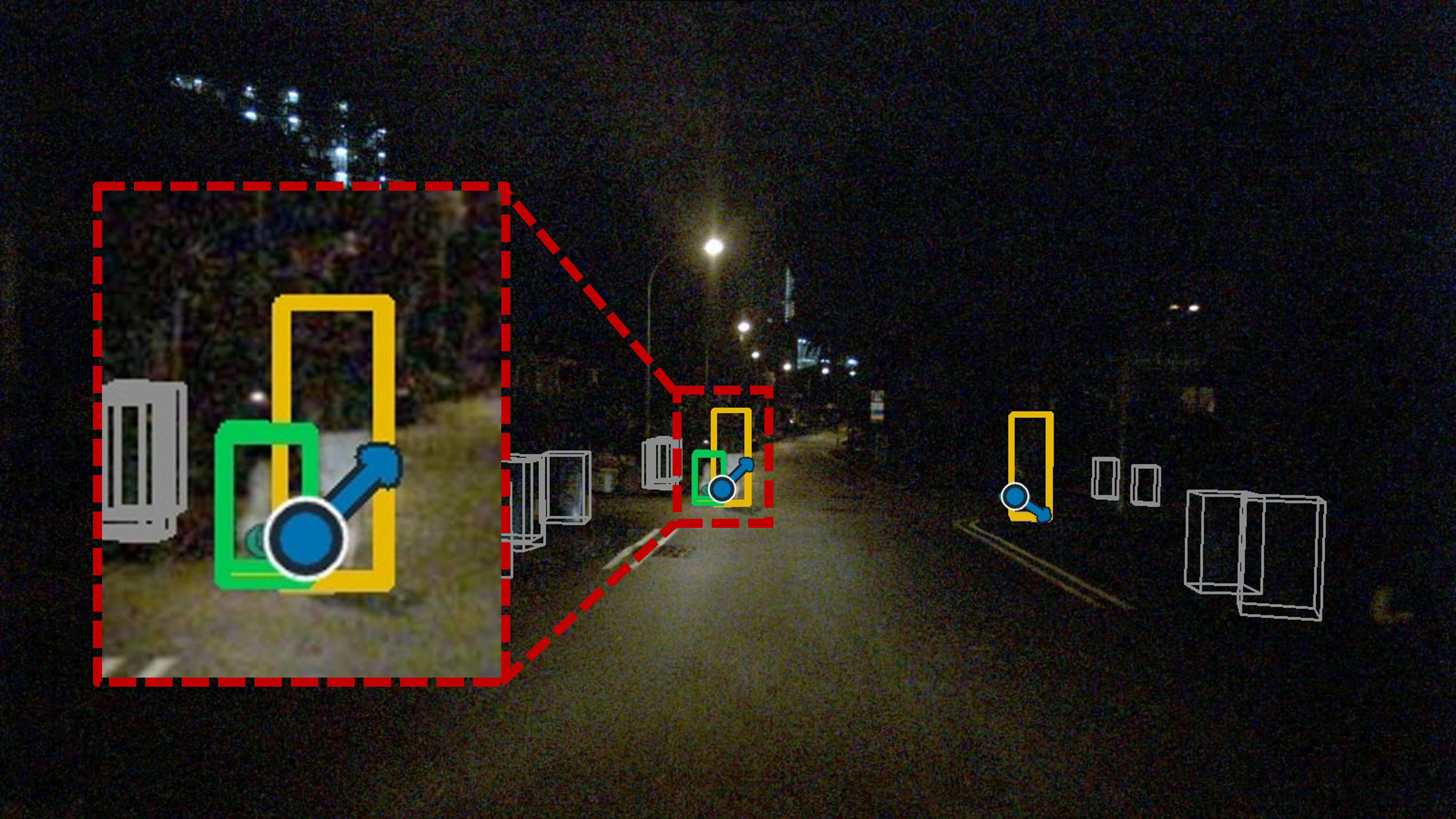}}
\\[-1.2ex]
\subfloat[Severely occluded crossing vehicle.]{
    \includegraphics[width=0.24\linewidth]{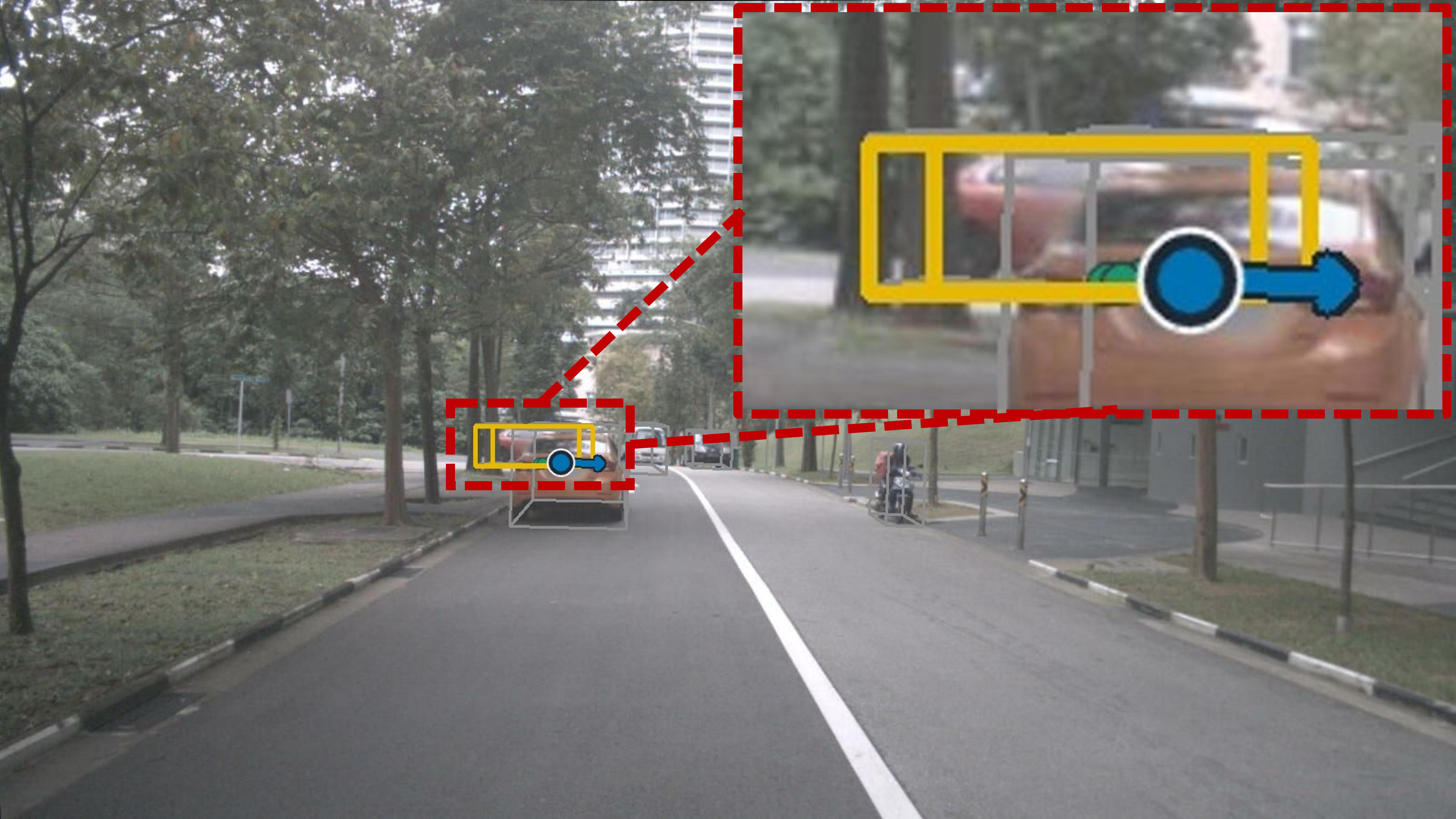}}
\hfill
\subfloat[Severely occluded moving vehicle.]{
    \includegraphics[width=0.24\linewidth]{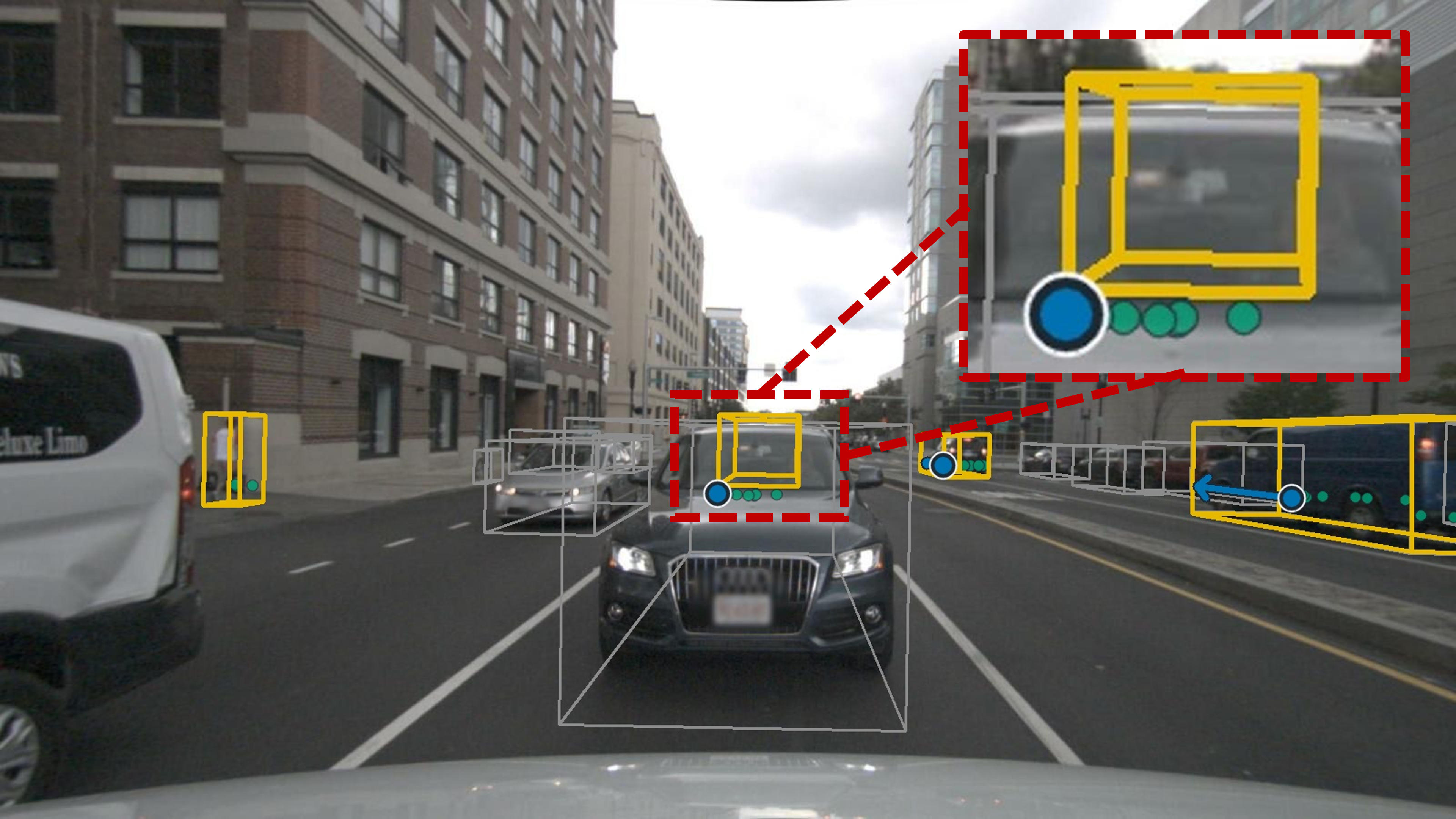}}
\hfill
\subfloat[Rainy-night vehicle under headlight glare.]{
    \includegraphics[width=0.24\linewidth]{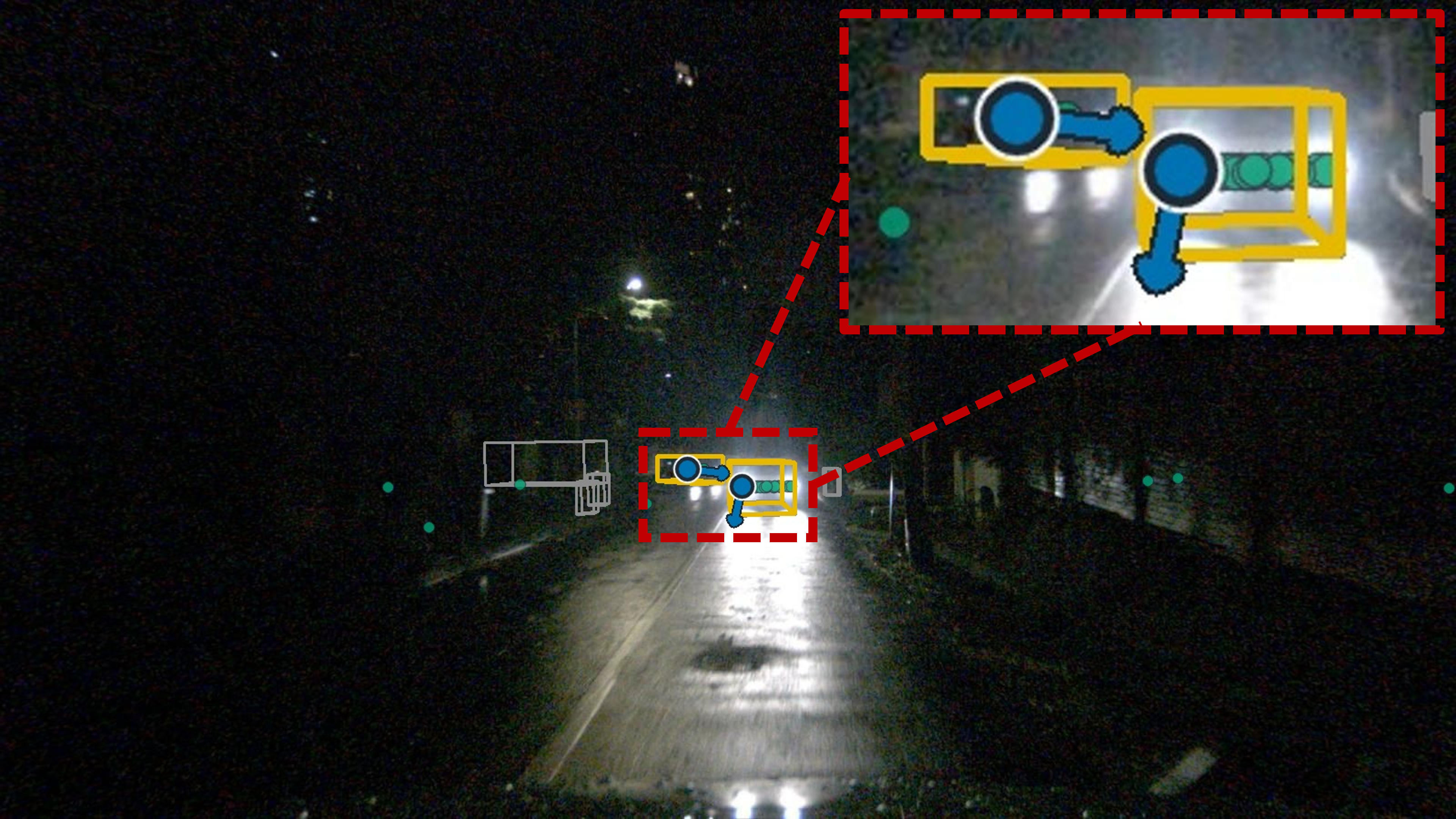}}
\hfill
\subfloat[Occluded crossing vehicle at night.]{
    \includegraphics[width=0.24\linewidth]{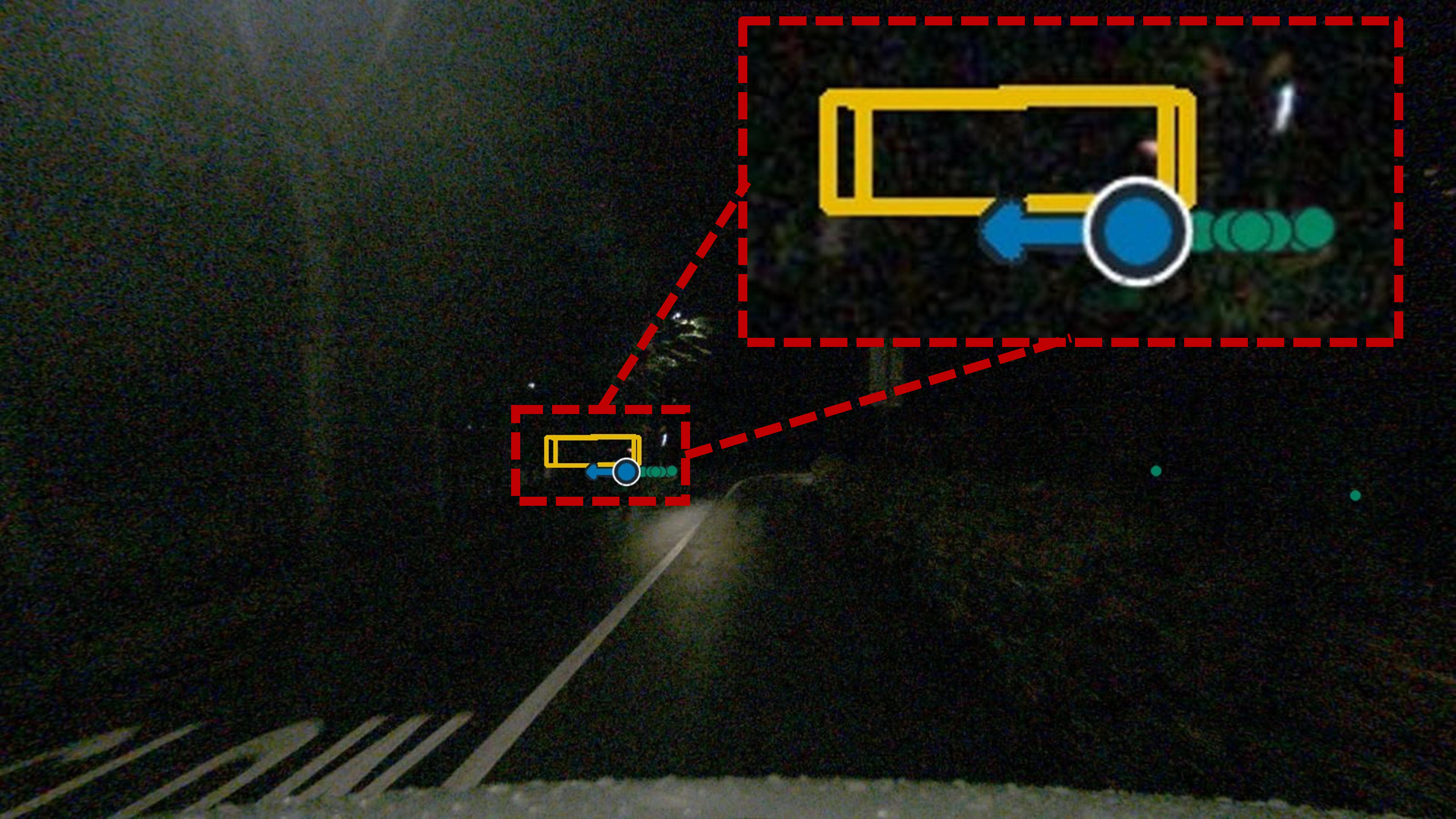}}
\\[-1.2ex]
\subfloat[UNANNOTATED vehicle in rain.]{
    \includegraphics[width=0.24\linewidth]{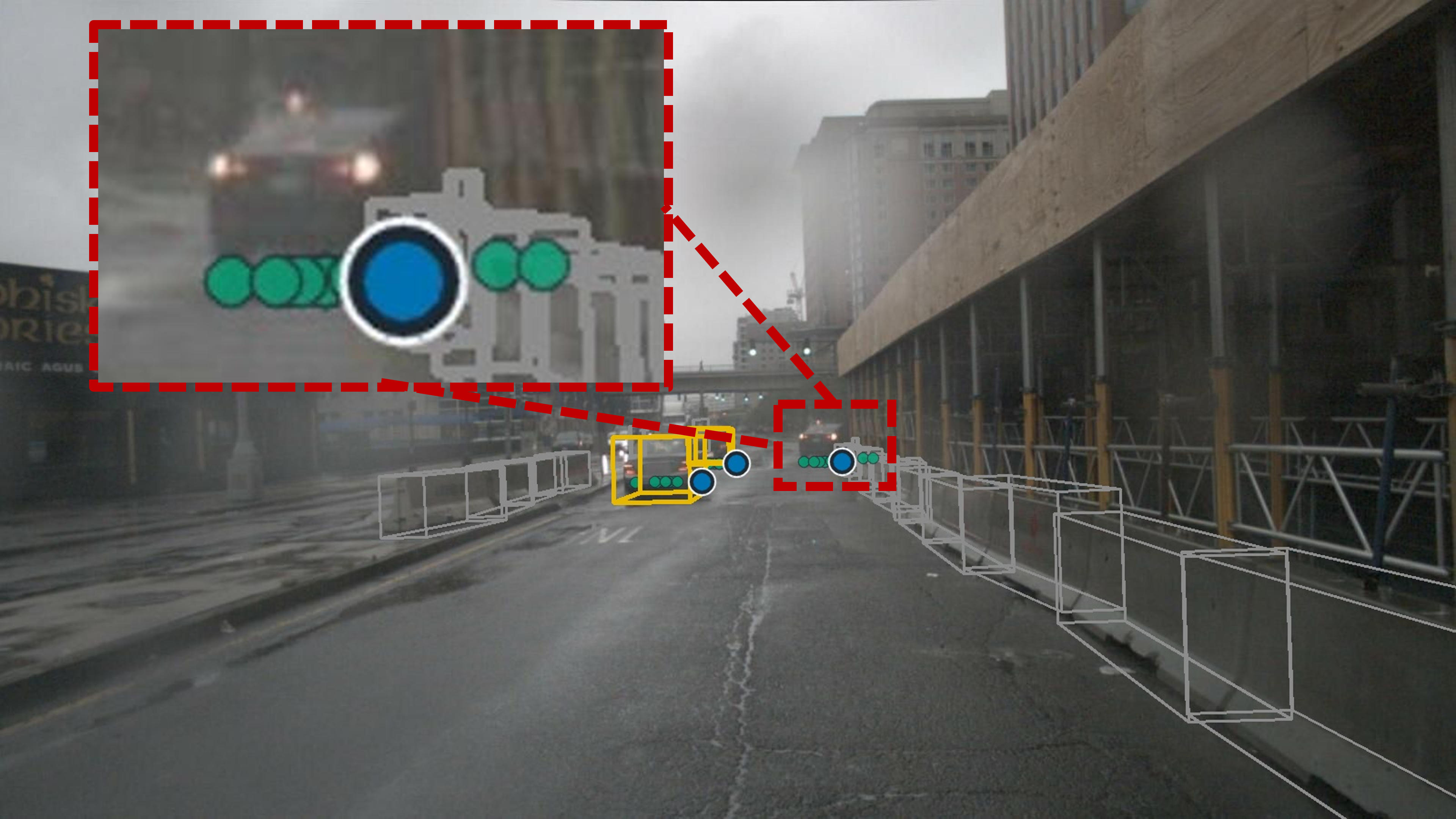}}
\hfill
\subfloat[UNANNOTATED vehicle under headlight glare.]{
    \includegraphics[width=0.24\linewidth]{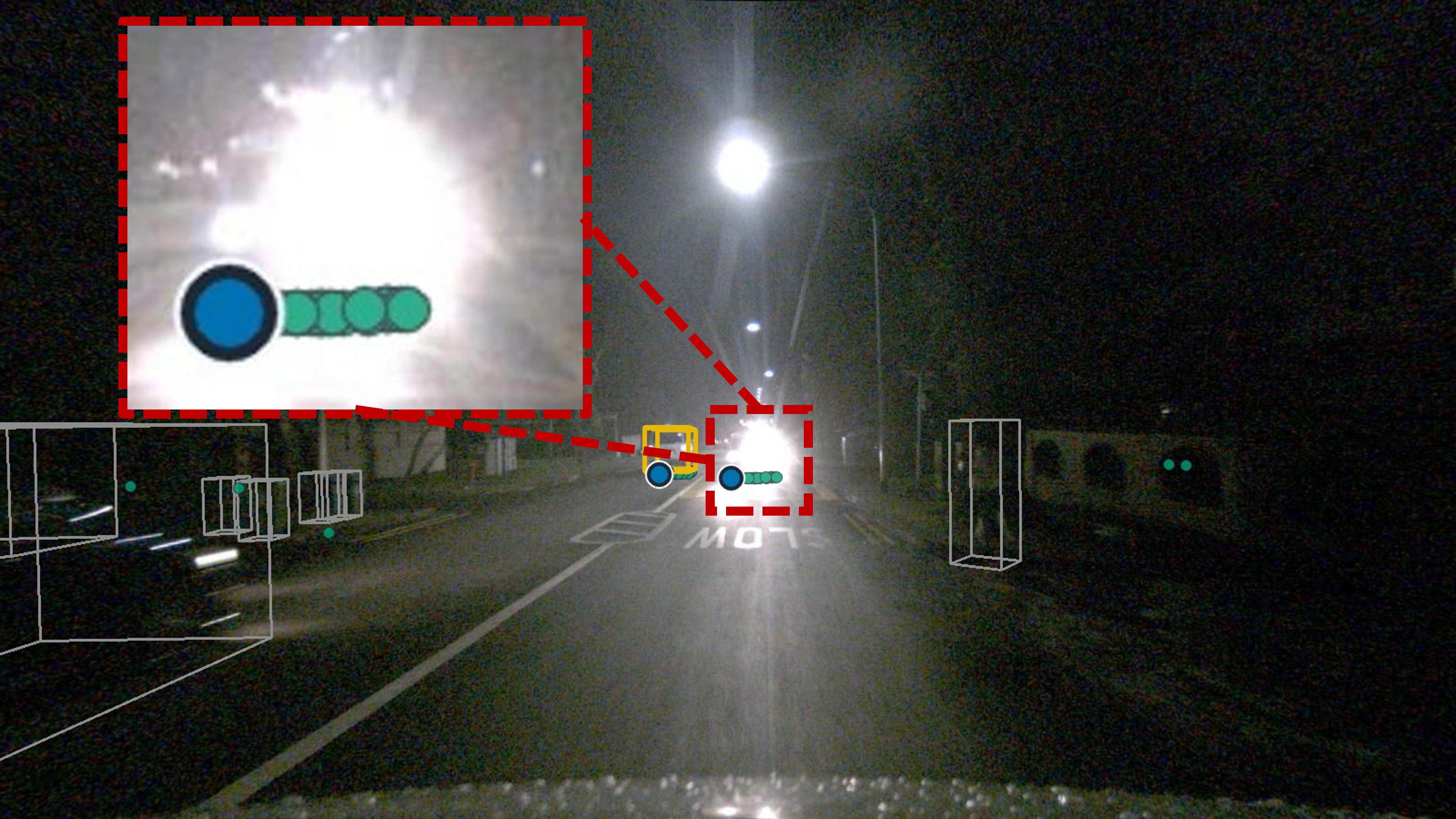}}
\hfill
\subfloat[UNANNOTATED oncoming vehicle in rain.]{
    \includegraphics[width=0.24\linewidth]{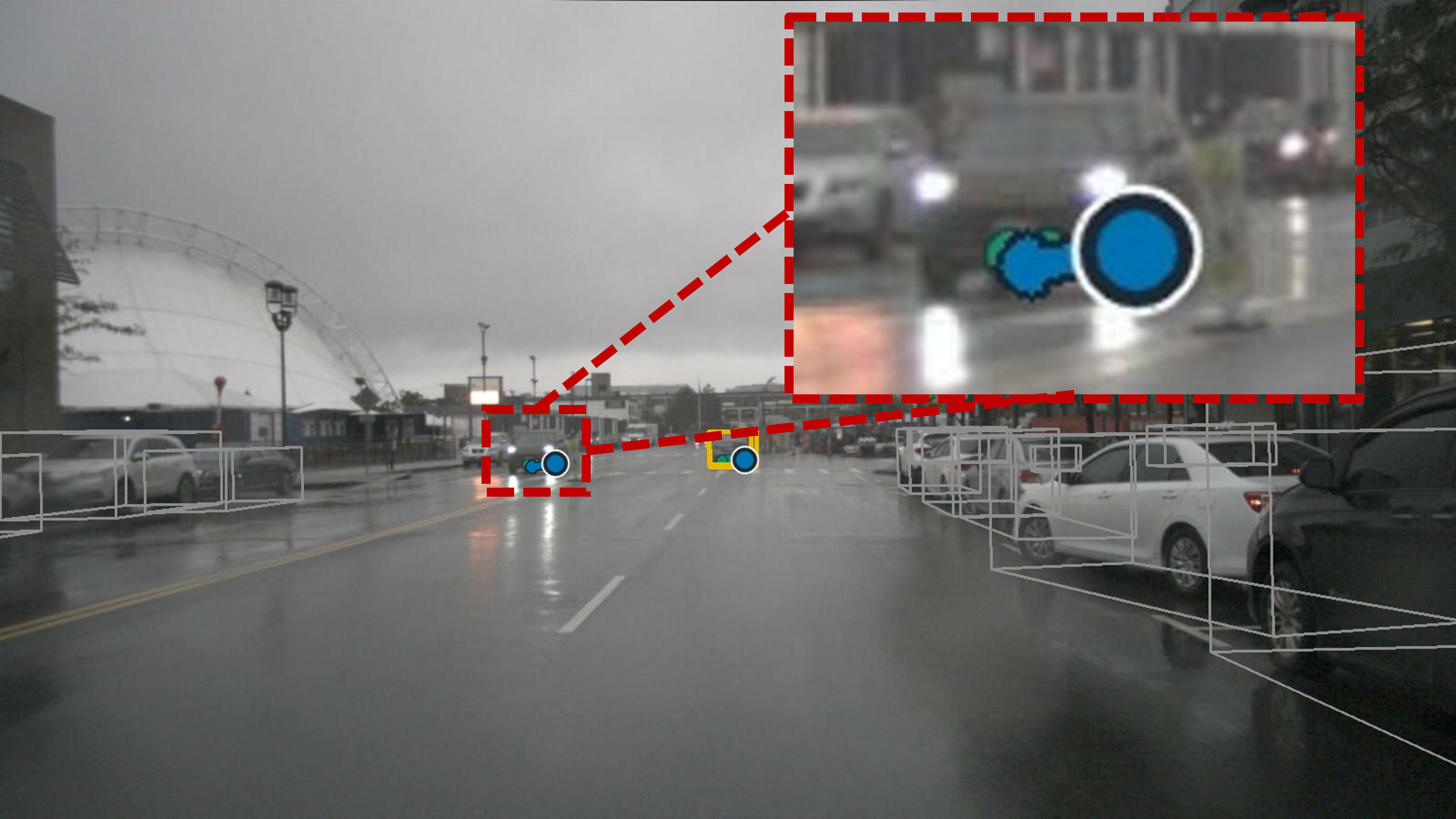}}
\hfill
\subfloat[UNANNOTATED motorcycle under headlight glare.]{
    \includegraphics[width=0.24\linewidth]{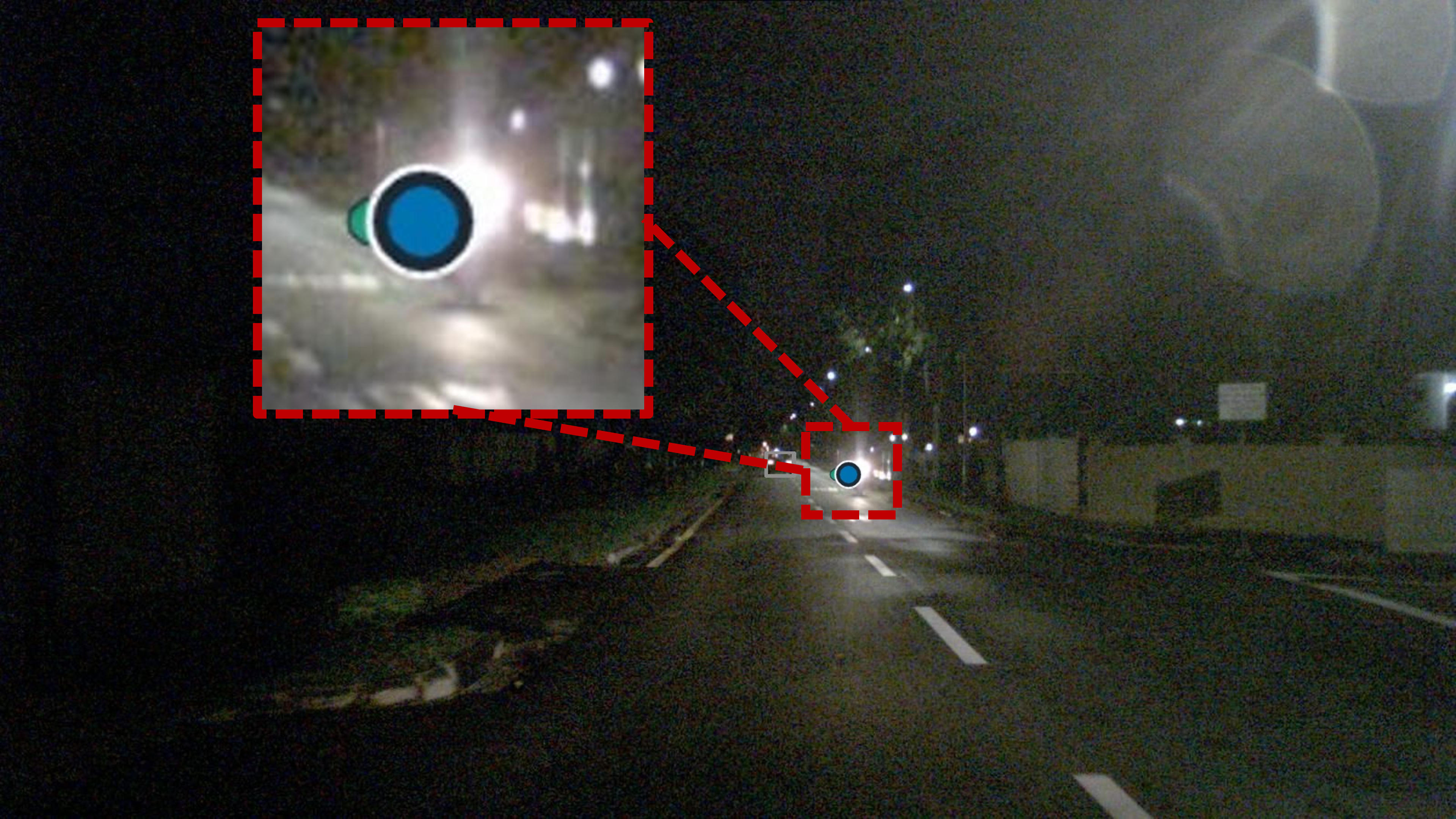}}

\caption{Qualitative results of PART in rare-object, adverse-condition, and
incomplete-annotation cases.
\textcolor[HTML]{00C853}{Bright-green},
\textcolor[HTML]{E6B800}{yellow}, and
\textcolor[HTML]{8C8C8C}{gray boxes} denote special-category moving objects
excluded from the standard nuScenes evaluation, moving ground truth used in
our evaluation, and other annotated objects, respectively.
\textcolor[HTML]{009E73}{Teal markers} show moving radar returns;
\textcolor[HTML]{0072B2}{blue markers and arrows} show PART predicted surface
points and ground-plane velocities, respectively.
(a)--(d) present rare and safety-relevant objects;
(e)--(h) cover occlusion, low illumination, rain, and headlight glare; and
(i)--(l) show moving objects unannotated in nuScenes. These examples illustrate PART’s behavior across rare objects, adverse conditions, and incomplete annotations.}
\label{fig:qualitative}
\end{figure*}

\section{Introduction}\label{sec:introduction}

Reliable environment perception remains essential for robotics and autonomous
driving. Even in emerging end-to-end and vision-language-action systems, object detection, semantic segmentation, and occupancy prediction
provide interpretable intermediate representations of the surrounding
scene~\cite{mao20233d,Chen10614862End2End}. Most existing detectors learn
visual appearance or lidar geometry from closed-set annotations. Their
reliability therefore depends on how well the training data cover the
deployment environment. Rare or unseen moving objects may be suppressed as
background, while adverse weather and unusual illumination can degrade sensor
observations~\cite{10377785Robo3D,10197287GLARE}. Additional data collection
and simulation can broaden this coverage, but rare events remain costly to
acquire and simulated measurements retain a gap from real sensors
~\cite{S2R-Bench}.

Unlike appearance, motion provides physical evidence shared across semantic
categories. Radar is well suited to exploiting this cue because it directly
measures radial velocity through the Doppler effect and is less sensitive to
illumination and adverse weather than cameras and lidar. It can therefore
indicate a moving object without first recognizing its appearance or semantic
category. However, sparse returns, limited angular resolution, sidelobe noise,
multipath reflections, and the lack of direct tangential-velocity measurements
make complete 3D box recovery difficult~\cite{Yang2020RadarNet}. For downstream
motion reasoning and collision avoidance, a representative surface location
and object velocity remain directly useful even when complete box geometry
cannot be reliably recovered. These outputs also align with radar sensing:
object returns provide evidence around reflecting surfaces, while Doppler
measurements constrain object motion. This motivates a class-agnostic moving-object formulation tailored to the
physical evidence provided by radar.

We propose the \textbf{P}hysics-\textbf{A}ware \textbf{R}adar
\textbf{T}ransformer (\textbf{PART}), a radar-only detector that grounds object
hypotheses in spatial--kinematic coherence, Doppler consistency, and radar
cross section (RCS). For each hypothesis, PART predicts an existence
confidence, a representative surface point, and a 2D ground-plane velocity.
Its uncertainty-aware supervision reduces reliance on exhaustive annotations
by preventing plausible radar-supported queries from being forced into the
negative class.

Our contributions are fourfold:

\begin{itemize}
    \setlength{\itemsep}{1pt}
    \setlength{\parskip}{0pt}
    \setlength{\parsep}{0pt}

    \item We develop PART, a fully sparse transformer for class-agnostic
    radar-only moving-object detection. It operates directly on sparse radar
    points without constructing dense bird's-eye-view (BEV) features. With only
    1.1 million parameters, PART provides a compact alternative to existing
    camera- and lidar-based detectors.

    \item We propose Doppler-Aware Query Initialization (DAQI), which forms
    input-dependent object queries from radar clusters coherent in position
    and velocity. Unlike scene-independent learned queries, these proposals
    are grounded in the current observations, reducing the burden of learning
    query--object correspondence from scratch.

    \item We introduce Physics-Guided Cross-Attention (PGCA), which constrains
    each proposed 2D velocity using the measured radial Doppler speeds.
    Point-level RCS and cluster-level RCS statistics are further incorporated
    into the attention projections, providing motion and scattering evidence
    for distinguishing object returns from radar noise and multipath
    reflections.

    \item We devise uncertainty-aware supervision to reduce reliance on
    exhaustive moving-object annotations. Random ground-truth masking and
    graded existence targets prevent plausible radar-supported queries from
    being treated as hard negatives, encouraging object existence to be
    inferred from physical radar evidence rather than annotation proximity.
\end{itemize}

Despite its compact design, PART achieves a class-agnostic average precision
(CA-AP) of $0.8827$, a mean average surface translation error (mASTE) of
$0.3188$~m, and a mean average velocity error (mAVE) of $0.8084$~m/s on
nuScenes~\cite{Caesar2019nuScenes}. It further attains $0.9203$ recall on rare and safety-relevant
categories excluded from the standard nuScenes evaluation and remains
effective under adverse weather, low illumination, and severe occlusion.
Fig.~\ref{fig:qualitative} presents representative detections, including
apparent false positives that correspond to moving objects absent from the
nuScenes annotations.

%%%%%%%%%%%%%%%%%%%%%%%%%%%%%%%%%%%%%%%%%%%%%%%%%%%%%%%%%%%%%%%%%%%%%%%%%%%%%%%%

\section{Related Work}\label{sec:related_work}

\subsection{Radar-Based 3D Object Detection}\label{subsec:radar_based_3d_object_detection}

Most 3D object detectors use lidar~\cite{Shi2022PVRCNN++,Yin2021CenterPoint,chen2023voxenext}, cameras~\cite{wang2021fcos3d,wang2021pgd,liu2022petr}, or their fusion~\cite{Liu2023BEVFusion,Song2024LiRaFusion,Kim2024CRTFusion}, owing to their accurate geometry, rich appearance, or both. Radar is usually an auxiliary modality that supplies range and Doppler cues~\cite{Song2024LiRaFusion,Kim2024CRTFusion}. Detection therefore remains dependent on lidar or camera when adverse weather or poor illumination degrades the primary input.

Radar-only 3D detection has received less attention because radar returns are sparse, noisy, and susceptible to multipath reflections. Existing methods adapt lidar-oriented designs using graph neural networks~\cite{Svenningsson2021RadarPointGNN} or hybrid grid--point backbones~\cite{Ulrich2022KPConvPillars}. AttentiveGRU~\cite{Saini2025AttentiveGRU} instead aggregates temporally correlated radar features, while RadarDistill~\cite{Geonho2024RadarDistill} transfers lidar features to a radar-only student.

\begin{figure*}[!t]
\centering
\includegraphics[width=0.75\linewidth]{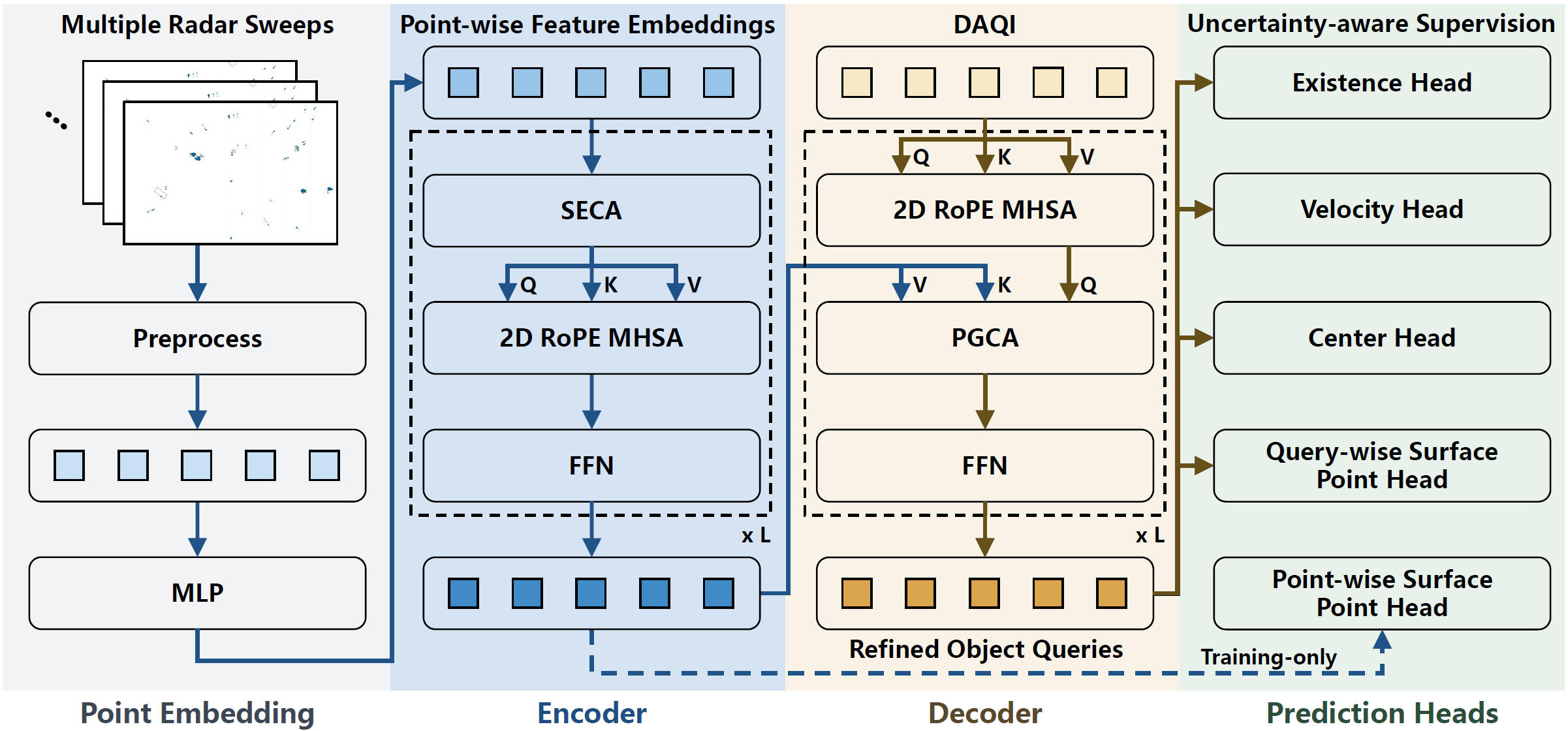}
\caption{Overview of the proposed PART. \textcolor[HTML]{205387}{Blue} paths encode radar points, while \textcolor[HTML]{665018}{gold} paths initialize and refine DAQI queries.}
\label{fig:overview}
\end{figure*}

These methods largely retain closed-set 3D box formulations inherited from lidar or depend on lidar supervision, leaving radar's Doppler geometry and RCS
evidence underused for target association. More fundamentally, a finite training taxonomy is unlikely to capture the full
diversity of moving objects encountered in real-world driving. Long-tail and previously unseen targets can produce coherent radar clusters,
while noise and multipath reflections may generate similar patterns.
Conventional closed-set supervision does not explicitly account for this
ambiguity, limiting generalization beyond the annotated categories.

\subsection{Transformer-Based 3D Detectors}
\label{subsec:transformer_based_3d_detectors}

DETR formulates detection as set prediction with learned queries and bipartite
matching, eliminating non-maximum suppression~\cite{Carion2020DETR}.
DETR3D extends this framework to multi-view 3D detection
~\cite{Wang2021DETR3D}. However, scene-independent queries provide limited
spatial guidance in sparse 3D scenes.

Lidar transformers adapt attention to sparse representations. Voxel
Transformer models interactions between voxels~\cite{Mao2021VoxelTransformer},
while SST avoids repeated downsampling~\cite{Fan2022SST}. CT3D refines
proposals from local point features~\cite{Shenga2021CT3D}, whereas CenterFormer
uses heatmap-derived center queries for spatial and temporal aggregation,
reducing query optimization difficulty~\cite{Zhou2022CenterFormer}. However,
global point attention has quadratic complexity and can be costly for dense
lidar inputs~\cite{Wu2024PointTransformer,Misra20213DETR}. These lidar-oriented
designs also omit radar-specific radial-Doppler constraints, while voxelization
may fragment the limited returns from one object.

%%%%%%%%%%%%%%%%%%%%%%%%%%%%%%%%%%%%%%%%%%%%%%%%%%%%%%%%%%%%%%%%%%%%%%%%%%%%%%%%

\section{PART}\label{sec:part}

As shown in Fig.~\ref{fig:overview}, PART processes temporally aggregated
radar points using a fully sparse encoder without voxelization or dense BEV
features. DAQI initializes object queries from position--velocity coherent
clusters, and stacked decoder layers refine them through query self-attention
and PGCA conditioned on radial Doppler and RCS information. The final queries
produce class-agnostic existence, surface-point, velocity, and auxiliary
center predictions. Training combines point-wise surface supervision with
uncertainty-aware existence targets.

\subsection{Point Embedding and Encoder}\label{subsec:point_embedding_and_encoder}

PART aggregates six sweeps from all five nuScenes radars, compensates for ego
motion, and transforms the returns to the current
\texttt{RADAR\_FRONT} frame. Returns with a compensated Doppler-speed
magnitude below $0.5$~m/s or outside the $\pm54$~m sensing region are removed. After motion filtering, the input contains fewer than 300 radar-point tokens in most nuScenes samples and never exceeds 800, making global point-wise attention computationally practical. Each retained point is represented by 16 features: 2D position, radar cross section (RCS), compensated velocity, six sensor-reported uncertainty terms, sweep timestamp, radial speed and direction, and a $k$-nearest-neighbor density estimate ($k=5$). Position and velocity support DAQI, while Doppler
and RCS provide physical inputs to PGCA. A three-layer MLP maps each vector to a 128-dimensional embedding.

The encoder stacks three transformer blocks. Each block applies SE-inspired
channel attention (SECA)~\cite{Hu2018SENet}, 2D RoPE multi-head self-attention
(MHSA)~\cite{Heo2024RoPEViT}, and a feed-forward network
(FFN). SECA uses a shared point-wise gating MLP to produce input-dependent
channel weights without sweep-level pooling. RoPE rotates queries and keys
using normalized 2D coordinates, thereby encoding relative displacement in
the attention scores while leaving values unchanged. Residual connections
and layer normalization follow MHSA and the FFN. The encoded point features
serve as keys and values for the decoder.

\subsection{Doppler-Aware Query Initialization (DAQI)}\label{subsec:daqi}

Standard DETR-style detectors use scene-independent learned queries whose correspondence to sparse radar observations must be established entirely during decoding. DAQI instead derives input-dependent queries by applying
DBSCAN~\cite{Ester1996DBSCAN} in the joint $[x,y,v_x,v_y]$ space to form
clusters of at least three retained radar points with similar positions
and ego-compensated velocities. For cluster $\mathcal C_k$, its centroid and pooled encoder feature initialize the query position and feature:
\begin{equation}
\mathbf p_k =
\frac{1}{|\mathcal C_k|}
\sum_{i\in\mathcal C_k}[x_i,y_i]^{T},
\qquad
\mathbf q_k^{\mathrm{init}} =
\frac{1}{|\mathcal C_k|}
\sum_{i\in\mathcal C_k}\mathbf f_i,
\label{eq:daqi}
\end{equation}
where $[x_i,y_i]^T$ and $\mathbf f_i$ denote the 2D position and encoded
feature of radar point $i$, respectively. We deliberately zero-initialize the final layers of the offset heads, so
that their initial position predictions coincide with the cluster centroid
$\mathbf p_k$.

We use a fixed budget of $K=100$ queries. During training, DAQI adds a balanced set of random queries away from DBSCAN proposals and annotated matching regions. At inference, only cluster-derived queries remain. These clusters are proposals rather than confirmed objects, since radar noise
and multipath returns may also form coherent groups. PGCA
(Section~\ref{subsec:pgca}) incorporates Doppler consistency and RCS evidence
to help distinguish object returns from such interference.

\subsection{Physics-Guided Cross-Attention (PGCA)}\label{subsec:pgca}

\begin{figure}[!t]
\centering
\includegraphics[width=0.8\linewidth]{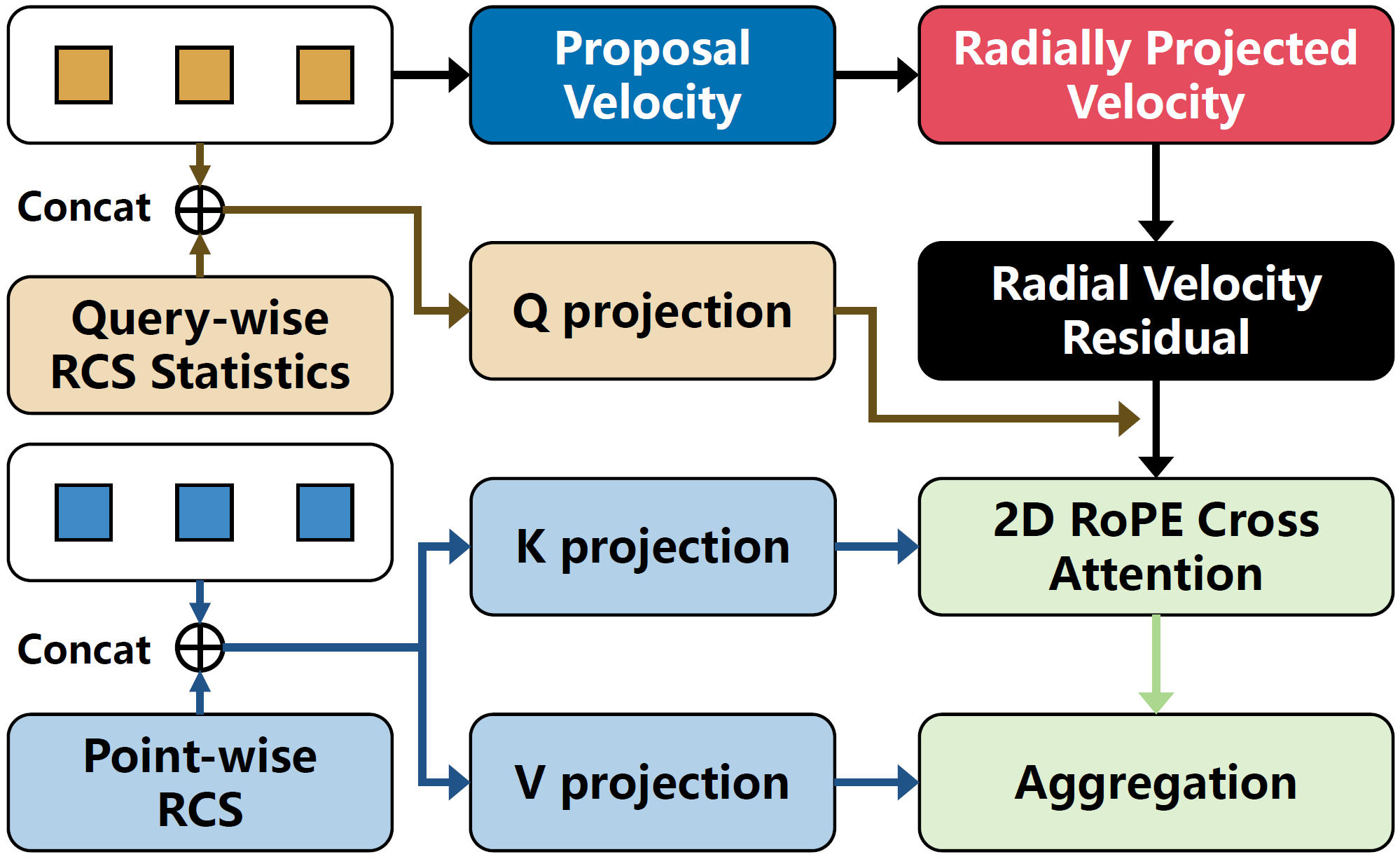}
\caption{Detailed architecture of PGCA.}
\label{fig:pgca_architecture}
\end{figure}

Each decoder layer first applies 2D RoPE MHSA to the active object queries.
Let $\mathbf q_k$ denote the resulting feature of query $k$. PGCA then
associates each query with the encoded radar points using feature, Doppler,
and RCS evidence. Its detailed architecture is shown in
Fig.~\ref{fig:pgca_architecture}.

\begin{figure}[!t]
\centering
\includegraphics[width=0.7\linewidth]{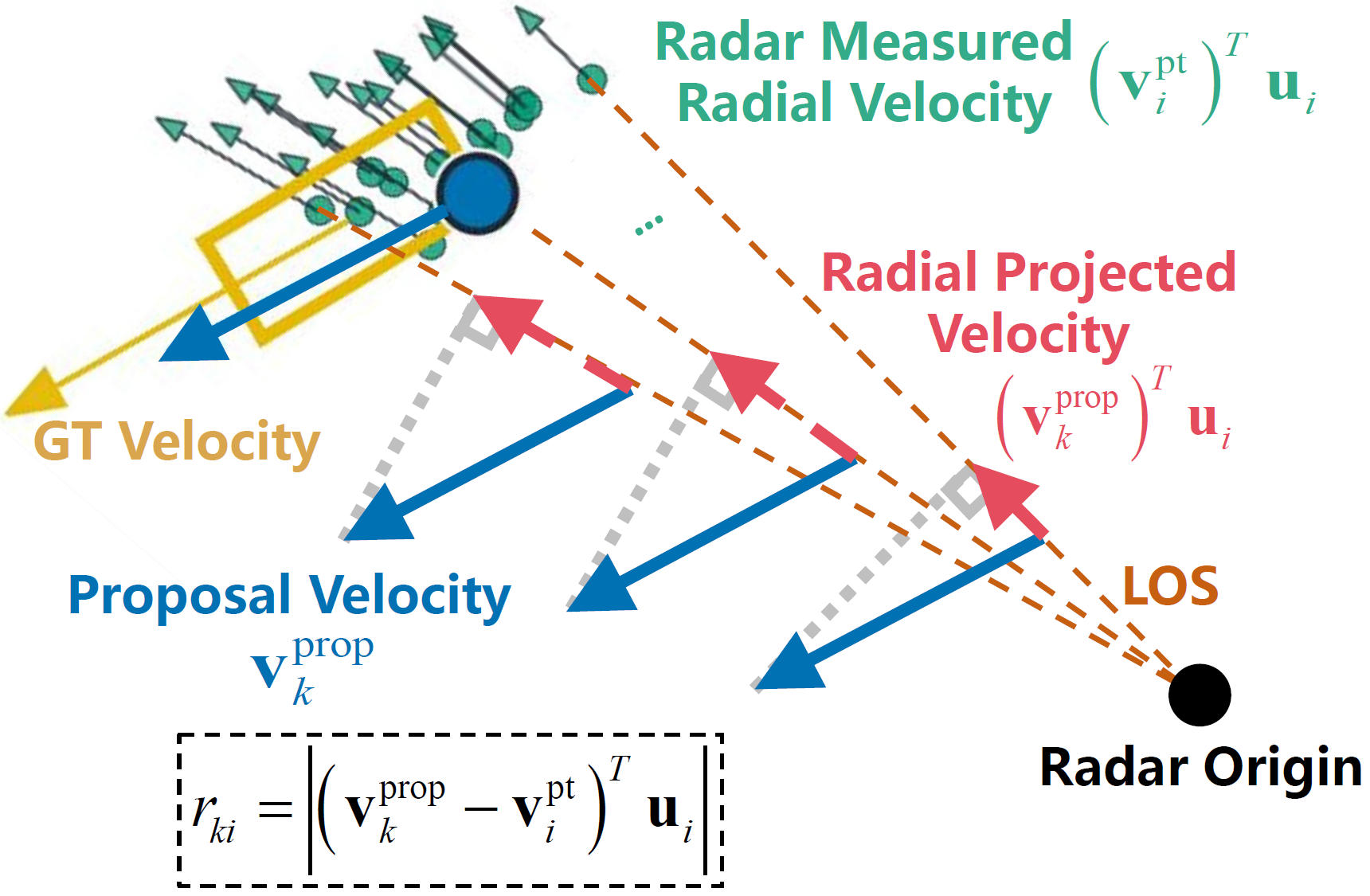}
\caption{Doppler-consistency geometry in PGCA. The 2D query velocity proposal
$\mathbf v_k^{\mathrm{prop}}$ and the ego-compensated point velocity
$\mathbf v_i^{\mathrm{pt}}$ are projected onto $\mathbf u_i$, the unit LOS direction from the source radar to point $i$ at its
measurement timestamp. All vectors are expressed in the reference frame.
The absolute difference between their signed LOS projections defines the
residual $r_{ki}$ used in cross-attention.}
\label{fig:doppler_projection}
\end{figure}

Radar measures only the signed component of motion along the sensor line of
sight (LOS). As illustrated in Fig.~\ref{fig:doppler_projection}, each radar
point provides motion evidence only along the radial direction, whereas a
query proposes a 2D ground-plane velocity. Let $\mathbf u_i$ denote the unit
LOS vector from the source radar to point $i$ at its measurement timestamp,
rotated into the reference frame, and let $\mathbf v_i^{\mathrm{pt}}$ denote
the ego-compensated ground-plane velocity of the point. For query $k$, PGCA
predicts a ground-plane velocity proposal and computes the discrepancy between
their signed radial components:
\begin{equation}
\mathbf v_k^{\mathrm{prop}} =
g_{\mathrm{prop}}(\mathbf q_k),
\qquad
r_{ki} =
\left|
\left(
\mathbf v_k^{\mathrm{prop}}-\mathbf v_i^{\mathrm{pt}}
\right)^{T}
\mathbf u_i
\right|.
\label{eq:doppler_residual}
\end{equation}
A large residual $r_{ki}$ indicates Doppler-inconsistent motion and reduces the association between query $k$ and radar point $i$.

For a cluster-derived query, we retain the cluster-wise RCS mean, standard
deviation, minimum, and maximum as
$\boldsymbol{\rho}_k =
[\mu_k,\sigma_k,c_k^{\min},c_k^{\max}]^{T}$.
The cluster statistics condition the query projection, while the point RCS
$c_i$ conditions the key and value projections:
\begin{equation}
\tilde{\mathbf q}_k=W_Q[\mathbf q_k;\boldsymbol{\rho}_k],\quad
\tilde{\mathbf k}_i=W_K[\mathbf f_i;c_i],\quad
\tilde{\mathbf v}_i=W_V[\mathbf f_i;c_i].
\label{eq:rcs_injection}
\end{equation}
This design allows PGCA to learn RCS compatibility from annotated radar
returns without imposing a fixed scattering rule.

Let $\mathcal R(\mathbf p)$ denote the 2D RoPE transformation at position
$\mathbf p$. For attention head $h$, the association weights are
\begin{equation}
\alpha_{ki}^{h}=
\operatorname{softmax}_{i}\!\left(
\frac{
[\mathcal R(\mathbf p_k)\tilde{\mathbf q}_k^{h}]^{T}
[\mathcal R(\mathbf p_i)\tilde{\mathbf k}_i^{h}]
}{\sqrt{d_h}}
-\lambda_v r_{ki}
\right),
\label{eq:pgca_attention}
\end{equation}
where $d_h$ is the head dimension and $\lambda_v$ is learnable. RCS enters
the feature similarity through the conditioned projections, while the
second term explicitly penalizes Doppler inconsistency.

The attended radar message is
\begin{equation}
\mathbf o_k^h=\sum_{i=1}^{N}\alpha_{ki}^h\tilde{\mathbf v}_i^h,
\qquad
\mathbf m_k=W_O
\left[\mathbf o_k^1\Vert\cdots\Vert\mathbf o_k^H\right],
\label{eq:pgca_message}
\end{equation}
where $N$ and $H$ are the numbers of radar points and attention heads,
respectively, $W_O$ is the output projection, and $\Vert$ denotes
channel-wise concatenation. Finally, $\mathbf m_k$ updates $\mathbf q_k$
through a residual connection and layer normalization, followed by an FFN
with a second residual update. The output is passed to the next decoder
layer, and the final-layer queries are decoded by the prediction heads.

\subsection{Prediction Heads and Loss}
\label{subsec:prediction-heads-and-loss}

After $L=3$ decoder layers, four query-wise heads decode an existence confidence, a surface point, a center, and a ground-plane velocity:
\begin{equation}
\begin{aligned}
\hat e_k &=
\sigma\!\left(g_{\mathrm{exist}}(\mathbf q_k^{(L)})\right),\\
\hat{\mathbf s}_k &=
\mathbf p_k+
g_{\mathrm{q\text{-}surf}}
\!\left([\mathbf q_k^{(L)};\mathbf p_k]\right),\\
\hat{\mathbf c}_k &=
\mathbf p_k+
g_{\mathrm{center}}
\!\left([\mathbf q_k^{(L)};\mathbf p_k]\right),\\
\hat{\mathbf v}_k &=
\mathbf v_k^{\mathrm{prop},(L)}
+g_{\Delta v}\!\left(\mathbf m_k^{(L)}\right).
\end{aligned}
\label{eq:prediction-heads}
\end{equation}
The existence prediction is class-agnostic and indicates whether a query
represents a moving object. The surface point is the primary localization
output because radar returns commonly originate from object surfaces. The
center head is retained for comparison under center-based protocols, while
the velocity head refines the final PGCA velocity proposal.

\begin{figure}[!t]
\centering
\includegraphics[width=0.8\linewidth]{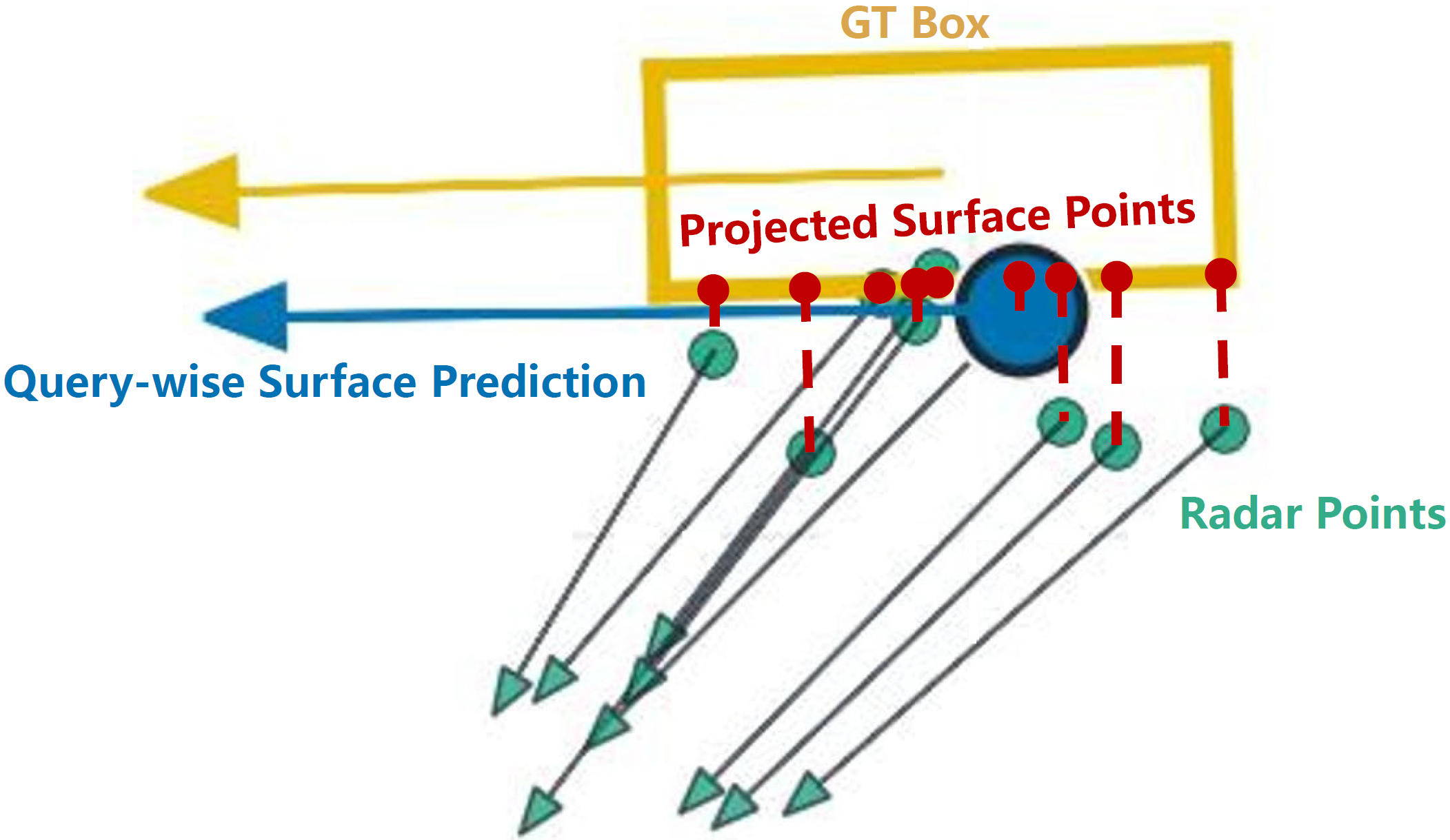}
\caption{Surface-point targets are obtained by projecting radar points and query references onto the nearest location on the ground-truth box perimeter.}
\label{fig:surface_point_projection}
\end{figure}

We match cluster-derived queries to moving ground-truth boxes using the
surface distance
\begin{equation}
d_{kj}=
\left\|
\hat{\mathbf s}_k-
\Pi_{\mathcal B_j}(\hat{\mathbf s}_k)
\right\|_2,
\label{eq:surface-cost}
\end{equation}
where $\Pi_{\mathcal B_j}(\cdot)$ projects a point onto the nearest location
on the ground-plane perimeter of box $\mathcal B_j$. As illustrated in
Fig.~\ref{fig:surface_point_projection}, the same geometric construction is
used to define both query-wise and point-wise surface targets. Each box supplies multiple matching slots.
Hungarian matching followed by a per-box top-$K$ rule yields the positive
query--box pairs $\mathcal P$; only DBSCAN-initialized queries can become
hard positives. For
$t\in\{\mathrm{q\text{-}surf},\mathrm{center},\mathrm{vel}\}$, the
regression losses share the form
\begin{equation}
\mathcal L_t=
\frac{1}{|\mathcal P|}
\sum_{(k,j)\in\mathcal P}
\left\|
\hat{\mathbf y}_k^t-\mathbf y_{kj}^{t*}
\right\|_1,
\label{eq:query-regression-loss}
\end{equation}
with targets
$\mathbf y_{kj}^{\mathrm{q\text{-}surf}*}
=\Pi_{\mathcal B_j}(\hat{\mathbf s}_k)$,
$\mathbf y_{kj}^{\mathrm{center}*}=\mathbf c_j$, and
$\mathbf y_{kj}^{\mathrm{vel}*}=\mathbf v_j$.

The final encoder block also includes an auxiliary point-wise surface head.
Before embedding, each radar point is associated with the most compatible
ground-truth box according to angular, surface, velocity-direction, and
radial-speed consistency. Let $\mathcal A$ denote the set of associated radar-point--box pairs
$(i,j)$. For each $(i,j)\in\mathcal A$,
\begin{equation}
\begin{aligned}
\hat{\mathbf s}_i^{\mathrm p}
&=\mathbf r_i+
g_{\mathrm{p\text{-}surf}}(\mathbf f_i),\\
\mathbf s_{i}^{\mathrm p*}
&=\Pi_{\mathcal B_j}(\mathbf r_i),\\
\mathcal L_{\mathrm{p\text{-}surf}}
&=\frac{1}{|\mathcal A|}
\sum_{\mathcal A}
\left\|
\hat{\mathbf s}_i^{\mathrm p}
-
\mathbf s_{i}^{\mathrm p*}
\right\|_1 ,
\end{aligned}
\label{eq:point-surface-loss}
\end{equation}
where $\mathbf r_i$ is the normalized 2D position of radar point $i$ in
the reference radar frame. Its predicted offset is embedded and added to the encoder feature, providing
local surface geometry for subsequent query decoding.

The complete objective is
\begin{equation}
\mathcal L=
\lambda_{\mathrm{exist}}\mathcal L_{\mathrm{exist}}
+\sum_{t}
\lambda_t\mathcal L_t
+\lambda_{\mathrm{p\text{-}surf}}
\mathcal L_{\mathrm{p\text{-}surf}},
\label{eq:total-loss}
\end{equation}
where the $\lambda$ terms are scalar loss weights and the summation covers
the three query-wise regression terms. The existence loss is defined by the
uncertainty-aware supervision described next.

\subsection{Uncertainty-Aware Supervision}
\label{subsec:uncertainty-aware-supervision}

The moving-object distribution in open-road driving has a long tail that
cannot be exhaustively represented by a fixed taxonomy or a finite training
set. Under conventional binary supervision, an unmatched radar-supported
query is treated as a hard negative, even when its motion pattern may
correspond to a rare or previously unseen object. To reduce this dependence
on annotation coverage, we independently mask each ground-truth box with
probability $r_{\mathrm{mask}}=0.5$ during training. This procedure exposes PART to radar-supported object patterns without hard
positive labels and encourages the existence head to rely on physical radar
evidence rather than the object distribution covered by the training set.

Let $\mathcal Q_{\mathrm p}$ contain DBSCAN queries matched to unmasked boxes within
the association threshold, and let $\mathcal Q_{\mathrm m}$ contain queries
matched to masked boxes. We further define $\mathcal Q_{\mathrm u}$ as
DBSCAN queries containing at least five points that remain unmatched
or fall outside the hard-positive threshold. Their existence targets are
\begin{equation}
y_k =
\begin{cases}
1,    & k\in\mathcal Q_{\mathrm p},\\
0.75, & k\in\mathcal Q_{\mathrm m},\\
0.5,  & k\in\mathcal Q_{\mathrm u},\\
0,    & \text{otherwise}.
\end{cases}
\label{eq:soft-existence-target}
\end{equation}
The target $0.75$ retains evidence for a deliberately masked object without
enabling regression, while $0.5$ preserves the ambiguity between an
unannotated object and radar clutter.

We optimize the existence head using sigmoid focal loss with
$\gamma=2.0$ and $\alpha=0.25$:
\begin{equation}
\mathcal L_{\mathrm{exist}} =
\frac{1}{\max(|\mathcal Q_{\mathrm p}|,1)}
\sum_{k\in\mathcal Q_{\mathrm t}}
\operatorname{FL}_{\gamma,\alpha}(z_k,y_k),
\label{eq:existence-loss}
\end{equation}
where $z_k=g_{\mathrm{exist}}(\mathbf q_k^{(L)})$ is the raw existence
logit and $\mathcal Q_{\mathrm t}$ is the complete set of supervised
queries. Only queries in $\mathcal Q_{\mathrm p}$ contribute to the query-wise
surface, center, and velocity losses.

%%%%%%%%%%%%%%%%%%%%%%%%%%%%%%%%%%%%%%%%%%%%%%%%%%%%%%%%%%%%%%%%%%%%%%%%%%%%%%%%
\section{Experiments}\label{sec:experiments}

\subsection{Experimental Protocol and Setup}
\label{subsec:experimental_and_setup}

\noindent\textbf{Dataset and protocol.}
We evaluate all methods on the official nuScenes validation split under a unified class-agnostic moving-object protocol. The evaluation region is limited to $\pm54$~m in the \texttt{RADAR\_FRONT} frame. Eligible ground-truth objects have a ground-plane speed above $0.5$~m/s; static targets are excluded.

\noindent\textbf{Metrics.}
Since PART is class-agnostic, all methods follow a unified class-agnostic
protocol. CA-AP is computed from the precision--recall curve. For matched true
positives, mASTE evaluates radar-facing surface localization: PART measures
the distance from its predicted surface point to the ground-truth box
perimeter, whereas box-based baselines measure the distance between the
predicted and ground-truth keypoints nearest the radar origin, each selected
from four BEV corners and four edge centers. The latter jointly captures
errors in box center, dimensions, and yaw. mATE and mAVE measure center and
ground-plane velocity errors, respectively.

\noindent\textbf{Implementation details.}
PART is trained for 30 epochs with Adam and a one-cycle schedule, using a batch size of eight per GPU, a peak learning rate of $1.2\times10^{-3}$, and a weight decay of $0.01$. Training uses GT sampling, flipping, rotation, and scaling. Sampled objects are radially perturbed, with up to $50\%$ of their returns dropped, while individual points receive radial Gaussian noise with $\sigma=0.5$~m. GT sampling and random GT masking are disabled during the final five epochs.

\subsection{Comparison with Existing 3D Detectors}\label{subsec:comparison_with_existing_3d_detectors}

\noindent\textbf{Overall comparison.}
Table~\ref{tab:overall} compares PART with representative detectors under the
unified class-agnostic protocol. With only 1.1M parameters, PART achieves a
CA-AP of $0.8827$, within $0.0050$ of the best result, and the lowest mASTE of
$0.3188$~m, outperforming the best fusion result by $0.0406$~m. Among
radar-input methods, PART improves RadarDistill by $0.1143$ CA-AP and reduces
mASTE by $58.9\%$, while using approximately $97\%$ fewer
parameters and requiring no lidar supervision. PART also achieves a
competitive mAVE of $0.8084$~m/s, compared with $0.7499$~m/s for RadarDistill. Compared with directly using DAQI proposals, PART improves
CA-AP by $0.2051$ and reduces mASTE and mAVE by $45.8\%$ and $48.7\%$,
respectively. These gains confirm that the learned encoder and physics-guided
decoder effectively refine the initial clustering proposals. Center
localization remains less accurate, reflecting the limited center evidence
available from sparse radar returns.

\begin{table}[!h]
\centering
\scriptsize
\setlength{\tabcolsep}{3pt}
\renewcommand\arraystretch{1.15}
\caption{Overall comparison with existing 3D detectors.}
\label{tab:overall}
\begin{threeparttable}
\begin{tabular}{lcccccc}
\hline\hline
\textbf{Methods} & \textbf{Params.}& \textbf{Mod.\tnote{1}} & \textbf{CA-AP$\uparrow$} & \textbf{mASTE$\downarrow$} & \textbf{mAVE$\downarrow$} & \textbf{mATE$\downarrow$} \\ \hline
CenterPoint~\cite{Yin2021CenterPoint} & 23M      & L & 0.8746                 & 0.4988               & \underline{0.7097}  & 0.4542                                                                           \\
VoxelNeXT~\cite{chen2023voxenext} & 31M        & L & 0.8801                 & 0.4498        & \textbf{0.6166}       & \underline{0.4172}                                                                           \\ 
TransFusion-L~\cite{Bai2022Transfusion} & 32M    & L & \underline{0.8865}                 & \underline{0.3617}        & 0.8600       & \underline{0.2790}                                                                           \\ \hline
FCOS3D~\cite{wang2021fcos3d} & 55M           & C & 0.5254                 & 1.2274         & 2.6908      & 1.2818                                                                           \\ 
PGD~\cite{wang2021pgd} & 56M              & C & 0.5423                 & 1.4796         & 2.7007      & 1.3143                                                                           \\ 
PETR~\cite{liu2022petr} & 83M             & C & 0.6553                 & 2.9833        & 2.3070       & 0.9340                                                                           \\ \hline
BEVFusion~\cite{Liu2023BEVFusion} & 157M        & LC & \textbf{0.8877}               & \underline{0.3594}        & 0.8138       & \textbf{0.2772}                                                                           \\
LiRaFusion~\cite{Song2024LiRaFusion} & 23M        & LR & \underline{0.8865}               & 0.5282       & 0.7556        & 0.4792                                                                           \\
CRTFusion~\cite{Kim2024CRTFusion}  & 81M        & RC & 0.8612               & 0.6699        & 0.7711       & 0.6206                                                                           \\ \hline
DAQI Proposals\tnote{2} & N/A           & R & 0.6776                 & 0.5884       & 1.5758        & 1.7806                                                                           \\
PillarNet-R~\cite{Shi2022PillarNet} & 16M      & R & 0.7492                 & 0.9586       & 0.8702        & 0.8919                                                                           \\ 
RadarDistill~\cite{Geonho2024RadarDistill} & 41M     & R(L)\tnote{3} & 0.7684                 & 0.7764        & \underline{0.7499}       & 0.7587                                                                           \\ \hline
\textbf{PART} & \textbf{1.1M}    & R & \underline{0.8827}                 & \textbf{0.3188}        & 0.8084       & 1.4846                                                                           \\ \hline\hline
\end{tabular}
\begin{tablenotes}
\item[] The best value is shown in \textbf{bold}; the next two best distinct
values are \underline{underlined}.
\item[1] L: lidar; C: camera; R: radar. 
\item[2] DAQI Proposals denotes the non-learned baseline that directly uses the DBSCAN cluster centroids and mean velocities as predictions.
\item[3] RadarDistill uses a lidar teacher during training but requires only radar input at inference.
\end{tablenotes}
\end{threeparttable}
\end{table}

\noindent\textbf{Robustness under Challenging Conditions.}
We identify the \emph{Night} and \emph{Rainy} subsets from keywords in the
official nuScenes scene descriptions; \emph{Night+Rainy} denotes their
intersection. As shown in Table~\ref{tab:challenging_conditions}, PART achieves
the highest CA-AP in all three subsets, exceeding the strongest competitor by
$1.14$, $0.52$, and $0.29$ percentage points, respectively. Relative to the best radar-input baseline, the margins widen to $7.68$,
$9.20$, and $6.26$ percentage points, respectively. PART also obtains the lowest mASTE in \emph{Rainy} and
\emph{Night+Rainy}, and the lowest mAVE in \emph{Night} and
\emph{Night+Rainy}; its nighttime mASTE is only $0.0055$m above the best
result. In \emph{Night+Rainy}, PART reduces mASTE by $32.4\%$ relative to the
next-best method and mAVE by $55.1\%$ relative to the best competing
radar-only inference method. These results demonstrate robust surface
localization and velocity estimation under low-light and rainy conditions.

\begin{table*}[htbp]
\centering
\footnotesize
\setlength{\tabcolsep}{3pt}
\renewcommand{\arraystretch}{1.15}
\caption{Robustness comparison under challenging conditions.}
\label{tab:challenging_conditions}
\begin{threeparttable}
\begin{tabular}{l c *{3}{c}@{\hspace{12pt}} *{3}{c}@{\hspace{12pt}} *{3}{c}}
\hline\hline
\multirow{2}{*}{\textbf{Method}} &
\multirow{2}{*}{\textbf{Mod.}} &
\multicolumn{3}{c}{\textbf{Night}} &
\multicolumn{3}{c}{\textbf{Rainy}} &
\multicolumn{3}{c}{\textbf{Night+Rainy}} \\
\cline{3-11}
& & \textbf{CA-AP$\uparrow$} & \textbf{mASTE$\downarrow$} & \textbf{mAVE$\downarrow$}
  & \textbf{CA-AP$\uparrow$} & \textbf{mASTE$\downarrow$} & \textbf{mAVE$\downarrow$}
  & \textbf{CA-AP$\uparrow$} & \textbf{mASTE$\downarrow$} & \textbf{mAVE$\downarrow$} \\
\hline
CenterPoint      & L  & 0.8297 & 0.3398 & 0.7795 & 0.8528 & 0.5671 & \underline{0.6729} & 0.8107 & 0.3410 & 0.7752 \\
VoxelNeXT        & L  & 0.8361 & 0.3602 & \underline{0.6968} & 0.8606 & 0.4503 & \textbf{0.5776} & 0.7861 & 0.4930 & \underline{0.6652} \\
TransFusion-L    & L  & 0.8584 & \textbf{0.2530} & \underline{0.6825} & 0.8662 & \underline{0.3699} & 0.8649 & 0.8520 & \underline{0.3122} & 0.6785 \\
\hline
FCOS3D           & C  & 0.5197 & 0.6753 & 2.0607 & 0.5203 & 1.3734 & 2.6823 & 0.4973 & 0.6429 & 2.3148 \\
PGD              & C  & 0.6035 & 0.6951 & 2.6314 & 0.5114 & 1.2124 & 2.5529 & 0.5704 & 0.5634 & 1.9414 \\
PETR             & C  & 0.6855 & 1.6695 & 2.3924 & 0.6978 & 3.2089 & 2.0686 & 0.7257 & 1.4301 & 2.5327 \\
\hline
BEVFusion        & LC & 0.8665 & \underline{0.2533} & 0.7634 & \underline{0.8754} & \underline{0.4137} & 0.8732 & 0.8768 & 0.3278 & 0.7924 \\
LiRaFusion        & LR & \underline{0.8816} & 0.3382 & 0.7499 & \underline{0.8864} & 0.5756 & 0.8066 & \underline{0.8793} & \underline{0.3007} & \underline{0.5332} \\
CRTFusion        & CR & \underline{0.8683} & 0.4730 & 0.8821 & 0.8480 & 0.7055 & 1.0037 & \underline{0.8863} & 0.3864 & 1.0296 \\
\hline
DAQI Proposals           & R  & 0.7026 & 0.4104 & 1.2325 & 0.6789 & 0.7148 & 1.6498 & 0.6889 & 0.4865 & 1.2728 \\
PillarNet-R      & R  & 0.8084 & 0.6602 & 0.9246 & 0.7711 & 1.0402 & 1.0874 & 0.8266 & 0.6334 & 0.6675 \\
RadarDistill     & R(L)  & 0.8162 & 0.3763 & 0.8175 & 0.7996 & 0.8528 & 0.7877 & 0.8164 & 0.4360 & 2.0693 \\
\hline
\textbf{PART}    & R  & \textbf{0.8930} & \underline{0.2585} & \textbf{0.4651}
                 & \textbf{0.8916} & \textbf{0.2811} & \underline{0.7590}
                 & \textbf{0.8892} & \textbf{0.2034} & \textbf{0.2994} \\
\hline\hline
\end{tabular}
\end{threeparttable}
\end{table*}

\begin{table*}[htbp]
\centering
\footnotesize
\setlength{\tabcolsep}{3pt}
\renewcommand{\arraystretch}{1.15}
\caption{Comparison on rare and safety-relevant object categories.}
\label{tab:rare_categories}
\begin{threeparttable}
\begin{tabular}{l c *{3}{c}@{\hspace{12pt}} *{3}{c}@{\hspace{12pt}} *{3}{c}@{\hspace{12pt}} *{3}{c}}
\hline\hline
\multirow{2}{*}{\textbf{Method}} &
\multirow{2}{*}{\textbf{Mod.}} &
\multicolumn{3}{c}{\textbf{Police Vehicles ($N_{\mathrm{GT}}=70$)}} &
\multicolumn{3}{c}{\textbf{Wheelchair ($N_{\mathrm{GT}}=260$)}} &
\multicolumn{3}{c}{\textbf{Personal Mobility ($N_{\mathrm{GT}}=102$)}} &
\multicolumn{3}{c}{\textbf{Animal ($N_{\mathrm{GT}}=13$)}} \\
\cline{3-14}
& & \textbf{Recall$\uparrow$} & \textbf{ASTE$\downarrow$} & \textbf{AVE$\downarrow$}
  & \textbf{Recall$\uparrow$} & \textbf{ASTE$\downarrow$} & \textbf{AVE$\downarrow$}
  & \textbf{Recall$\uparrow$} & \textbf{ASTE$\downarrow$} & \textbf{AVE$\downarrow$}
  & \textbf{Recall$\uparrow$} & \textbf{ASTE$\downarrow$} & \textbf{AVE$\downarrow$} \\
\hline
CenterPoint      & L  & 0.8286 & 0.2633 & 0.5565 & 0.0346 & 0.4032 & 0.3574 & 0.1863 & 0.2331 & 0.4617 & 0.1538 & 0.0979 & 0.2048 \\
VoxelNeXT        & L  & 0.8571 & 0.2872 & 0.6912 & 0.0115 & 0.3614 & 0.4953 & 0.1765 & 0.1923 & 0.3328 & \cellcolor{black!12}0.0000 & \cellcolor{black!12}{N/A} & \cellcolor{black!12}{N/A} \\
TransFusion-L    & L  & 0.5857 & 0.2038 & 0.6645 & 0.0038 & 0.4830 & 0.1431 & 0.0196 & 0.1621 & 0.4392 & \cellcolor{black!12}0.0000 & \cellcolor{black!12}{N/A} & \cellcolor{black!12}{N/A} \\
\hline
FCOS3D           & C  & 0.5286 & 0.5357 & 3.9904 & \cellcolor{black!12}0.0000 & \cellcolor{black!12}{N/A} & \cellcolor{black!12}{N/A} & 0.0588 & 0.2053 & 3.3621 & \cellcolor{black!12}0.0000 & \cellcolor{black!12}{N/A} & \cellcolor{black!12}{N/A} \\
PGD              & C  & 0.6143 & 0.4390 & 3.8758 & 0.0038 & 0.2099 & 1.5279 & 0.0392 & 0.1553 & 3.1449 & \cellcolor{black!12}0.0000 & \cellcolor{black!12}{N/A} & \cellcolor{black!12}{N/A} \\
PETR             & C  & \cellcolor{black!12}0.0000 & \cellcolor{black!12}{N/A} & \cellcolor{black!12}{N/A} & \cellcolor{black!12}0.0000 & \cellcolor{black!12}{N/A} & \cellcolor{black!12}{N/A} &  \cellcolor{black!12}0.0000 & \cellcolor{black!12}{N/A} & \cellcolor{black!12}{N/A} & \cellcolor{black!12}0.0000 & \cellcolor{black!12}{N/A} & \cellcolor{black!12}{N/A} \\
\hline
BEVFusion        & LC & 0.6000 & 0.3038 & 0.6558 & \cellcolor{black!12}0.0000 & \cellcolor{black!12}{N/A} & \cellcolor{black!12}{N/A} & \cellcolor{black!12}0.0000 & \cellcolor{black!12}{N/A} & \cellcolor{black!12}{N/A} & \cellcolor{black!12}0.0000 & \cellcolor{black!12}{N/A} & \cellcolor{black!12}{N/A} \\
LiRaFusion        & LR & 0.5143 & 0.2723 & 0.7073 & 0.0077 & 0.1715 & 0.2147 & \cellcolor{black!12}0.0000 & \cellcolor{black!12}{N/A} & \cellcolor{black!12}{N/A} & \cellcolor{black!12}0.0000 & \cellcolor{black!12}{N/A} & \cellcolor{black!12}{N/A} \\
CRTFusion        & CR & 0.6286 & 0.5582 & 0.7085 & 0.0192 & 0.2666 & 0.4843 & 0.1471 & 0.1546 & 1.6308 & \cellcolor{black!12}0.0000 & \cellcolor{black!12}{N/A} & \cellcolor{black!12}{N/A} \\
\hline
DAQI Proposals           & R  & 0.8571 & 0.5975 & 2.3529 & 0.8808 & 0.3043 & 0.3735 & 0.4706 & 0.1331 & 1.3884 & 0.6154 & 0.1305 & 0.1662 \\
PillarNet-R      & R  & 0.5588 & 0.4506 & 0.9424 & 0.0115 & 0.1633 & 0.2811 & \cellcolor{black!12}0.0000 & \cellcolor{black!12}{N/A} & \cellcolor{black!12}{N/A} & \cellcolor{black!12}0.0000 & \cellcolor{black!12}{N/A} & \cellcolor{black!12}{N/A} \\
RadarDistill     & R(L)  & 0.7647 & 0.3795 & 0.7877 & 0.0154 & 0.2340 & 0.4015 & \cellcolor{black!12}0.0000 & \cellcolor{black!12}{N/A} & \cellcolor{black!12}{N/A} & \cellcolor{black!12}0.0000 & \cellcolor{black!12}{N/A} & \cellcolor{black!12}{N/A} \\
\hline
\textbf{PART}    & R  & \textbf{0.9571} & 0.2931 & 1.0003
                 & \textbf{0.9577} & 0.1142 & 0.2161
                 & \textbf{0.8039} & 0.1063 & 0.6048 & \textbf{0.9231} & 0.2266 & 0.1838 \\
\hline\hline
\end{tabular}
\begin{tablenotes}
\item[] $N_{\mathrm{GT}}$ denotes the number of eligible ground-truth boxes in each
category. \textcolor[HTML]{5C5C5C}{Gray cells} indicate zero recall despite the presence of ground-truth instances. N/A denotes an undefined error because no true positive is available.
\end{tablenotes}
\end{threeparttable}
\end{table*}

\noindent\textbf{Rare and safety-relevant object categories.}
The standard nuScenes benchmark excludes six categories relevant to
moving-object detection: animals, wheelchairs, strollers, personal-mobility
users, ambulances, and police vehicles. Due to space constraints,
Table~\ref{tab:rare_categories} reports four representative categories.
Because eligible instances are scarce, we pool the official training and
validation splits for this diagnostic analysis. For this table only, PART is retrained with the same standard nuScenes
taxonomy as the baselines, without using ground-truth annotations from the
excluded categories during training. This simulates real-world conditions
in which long-tail moving objects lack annotations. Under
class-agnostic matching, category-specific false positives are undefined; we
therefore report recall and conditional errors.

PART achieves recalls of $95.71\%$, $95.77\%$, $80.39\%$, and $92.31\%$ for
police vehicles, wheelchair users, personal-mobility users, and animals,
surpassing the strongest alternatives by $10.00$, $7.69$, $33.33$, and $30.77$
percentage points, respectively. Some lidar- and camera-based detectors retain reasonable recall on
geometrically car-like police vehicles, but the gap to PART widens for the
other categories, where several learned baselines detect no personal-mobility
users or animals. Compared with raw DAQI proposals, PART raises recall from
$47.06\%$ to $80.39\%$ for personal mobility and from $61.54\%$ to $92.31\%$
for animals. Conditional ASTE and AVE are not ranked because the large recall gaps across
methods result in substantially different detected subsets. Nevertheless, PART attains ASTEs of
$0.1142$~m and $0.1063$~m for wheelchair and personal-mobility users,
respectively.

\subsection{Ablation and Additional Analyses}\label{subsec:ablation_study}

\noindent\textbf{Component ablation.}
Table~\ref{tab:ablation} evaluates the main components of PART. Removing DAQI
causes the largest degradation: CA-AP decreases by $0.2629$, while mASTE and
mAVE increase by $0.3589$~m and $0.4964$~m/s, respectively, even with the
relaxed score threshold. Special-category recall also falls from $0.9203$
to $0.6646$. Removing PGCA degrades all metrics, supporting the
joint use of radial-Doppler consistency and RCS information in cross-attention.
Without uncertainty-aware supervision, CA-AP drops to $0.7948$ and
special-category recall falls sharply to $0.3899$, confirming its importance
for generalization beyond the training taxonomy. Finally, removing PWSH leaves CA-AP unchanged but increases mASTE and mAVE
and reduces special-category recall, supporting the benefits of auxiliary
point-wise geometric supervision.

\begin{table}[!h]
\centering
\renewcommand{\arraystretch}{1.15}
\caption{Component ablation.}
\label{tab:ablation}
\begin{threeparttable}
\begin{tabular}{ccccc}
\hline\hline
\textbf{Exp.}                                  & \textbf{CA-AP$\uparrow$}       & \textbf{mASTE$\downarrow$}       & \textbf{mAVE$\downarrow$}           & \textbf{Special Recall$\uparrow$}\tnote{4}      \\ \hline
w/o DAQI\tnote{1}                          & 0.6198                     & 0.6777                     & 1.3048                        & 0.6646                     \\ 
w/o PGCA                          & \underline{0.8815}                     & \underline{0.3300}                     & 0.8348 & \underline{0.8931}                     \\ 
w/o UAS\tnote{2} & \underline{0.7948} & \textbf{0.3149} & \underline{0.8292} & 0.3899 \\
w/o PWSH\tnote{3}       & \textbf{0.8827}                     & 0.3607                     & \underline{0.8325}                        & \underline{0.9182}                     \\ \hline
\textbf{FULL}                              & \textbf{0.8827}            & \underline{0.3188}            & \textbf{0.8084}               & \textbf{0.9203}            \\ \hline\hline
\end{tabular}
\begin{tablenotes}
    \item[1] The w/o DAQI variant produces no valid detections at the
    standard score threshold $0.5$; its fixed-threshold metrics are
    therefore reported at $0.2$. CA-AP is computed over the full
    confidence sweep.
    \item[2] UAS: Uncertainty-Aware Supervision.
    \item[3] PWSH: Point-Wise Surface Head.
    \item[4] Special recall is evaluated over the rare and safety-relevant
    object categories.
\end{tablenotes}
\end{threeparttable}
\end{table}

\noindent\textbf{Unannotated moving objects.}
During false-positive analysis, we observed that some predictions
corresponded to visually identifiable moving objects not annotated in
nuScenes. Representative examples are shown in
Fig.~\ref{fig:qualitative}(i)--(l). All quantitative metrics use the unmodified nuScenes annotations; visually apparent but unannotated moving objects therefore remain counted as false positives. Although this observation is qualitative, it suggests that
PART can respond to motion-consistent radar evidence beyond the annotated
target set, rather than relying solely on the object distribution represented
by the training labels.

%%%%%%%%%%%%%%%%%%%%%%%%%%%%%%%%%%%%%%%%%%%%%%%%%%%%%%%%%%%%%%%%%%%%%%%%%%%%%%%%
\section{Conclusion}\label{sec:conclusion}

This paper presented PART, a lightweight radar-only transformer for
class-agnostic moving-object detection. Instead of predicting semantic
categories or complete 3D boxes, PART estimates an existence confidence,
a representative surface point, and ground-plane velocity from sparse
radar returns. DAQI grounds object queries in spatially and kinematically
coherent clusters, while PGCA incorporates radial-Doppler consistency and
RCS information into cross-attention. Uncertainty-aware supervision
further reduces dependence on exhaustive moving-object annotations. With
only 1.1 million parameters, PART achieves a CA-AP of $0.8827$, an mASTE
of $0.3188$~m, and an mAVE of $0.8084$~m/s on nuScenes. The results also
show strong recall for rare and safety-relevant objects and stable
performance under nighttime, rain, and severe occlusion. Qualitative
analysis further identifies valid detections of moving objects that are
unannotated in the dataset.

PART currently requires sufficient radar support and measurable radial
motion. Static objects and objects moving mainly in the tangential
direction remain outside its present scope, as do semantic classification
and complete 3D box estimation. Future work will extend PART toward temporal tracking and broader integration with general-purpose perception systems while retaining radar-only operation.

%%%%%%%%%%%%%%%%%%%%%%%%%%%%%%%%%%%%%%%%%%%%%%%%%%%%%%%%%%%%%%%%%%%%%%%%%%%%%%%%
% TODO: CAMERA-READY ONLY: remove this section for the double-anonymous submission.
\section*{Acknowledgment}\label{sec:acknowledgment}

This work was supported in part by the China Postdoctoral Science Foundation under Grant 2026M791817, by the National Natural Science Foundation of China under Grants 62533006 and 52471376, and by the Center for HPC, University of Electronic Science and Technology of China.

%%%%%%%%%%%%%%%%%%%%%%%%%%%%%%%%%%%%%%%%%%%%%%%%%%%%%%%%%%%%%%%%%%%%%%%%%%%%%%%%
% References.
% NOTE: several entries are best-effort scaffold placeholders. Fill full author
% lists and verify DOIs/arXiv IDs during the citation-compliance pass before
% submission. Do NOT submit with incomplete author fields.

\bibliographystyle{IEEEtran} 
\bibliography{root}

\end{document}